\documentclass[11pt]{article}

\usepackage[utf8]{inputenc}
\usepackage[margin=1in]{geometry}
\usepackage{times}
\usepackage{natbib}
\usepackage{graphicx}
\usepackage{amsmath}
\usepackage{amssymb}
\usepackage{url}
\usepackage{hyperref}
\usepackage{subcaption}
\usepackage{gensymb}
\hypersetup{colorlinks=true,linkcolor=blue,citecolor=blue,urlcolor=blue}

\title{
\textbf{Composite Safety Potential Field for Highway Driving Risk Assessment} \\[1em]
\normalsize A PREPRINT
}
\author{
Dachuan Zuo \quad Zilin Bian$^{*}$ \quad Fan Zuo \quad Kaan Ozbay \\
\normalsize New York University \\
\normalsize\texttt{\{dz2231,zb536,fz380,kaan.ozbay\}@nyu.edu}
}
\date{}  

\begin{document}

\maketitle

\begin{abstract}
In the era of rapid advancements in vehicle safety technologies, driving risk assessment has become a focal point of attention. Technologies such as collision warning systems, advanced driver assistance systems (ADAS), and autonomous driving require driving risks to be evaluated proactively and in real time. To be effective, driving risk assessment metrics must not only accurately identify potential collisions but also exhibit human-like reasoning to enable safe and seamless interactions between vehicles. Existing safety potential field models assess driving risks by considering both objective and subjective safety factors. However, their practical applicability in real-world risk assessment tasks is limited. These models are often challenging to calibrate due to the arbitrary nature of their structures, and calibration can be inefficient because of the scarcity of accident statistics. Additionally, they struggle to generalize across both longitudinal and lateral risks. To address these challenges, we propose a composite safety potential field framework, namely C-SPF, involving a subjective field to capture drivers' risk perception about spatial proximity and an objective field to quantify the imminent collision probability, to comprehensively evaluate driving risks. Different from existing models, the C-SPF is calibrated using abundant two-dimensional spacing data from trajectory datasets, enabling it to effectively capture drivers' proximity risk perception and provide a more realistic explanation of driving behaviors. Analysis of a naturalistic driving dataset demonstrates that the C-SPF can capture both longitudinal and lateral risks that trigger drivers' safety maneuvers. Further case studies highlight the C-SPF's ability to explain lateral driver behaviors, such as abandoning lane changes or adjusting lateral position relative to adjacent vehicles, which are capabilities that existing models fail to achieve.
\end{abstract}

\section{Introduction}\label{}

Driving risk assessment is a critical aspect of traffic safety research. It involves identifying factors that could compromise a vehicle's safety state and quantifying its proximity to potentially hazardous situations, such as collisions. In an era of rapidly emerging vehicle safety technologies, proactively assessing driving risks and delivering real-time inputs to safety algorithms are critical for the effectiveness of advanced safety systems, including collision warning systems, advanced driver assistance systems (ADAS), and autonomous driving solutions.

Numerous approaches have been developed for driving risk assessment. One category focuses on objective data, such as vehicle trajectories and underlying kinematics. Among the most widely used methods in this category are conflict-based surrogate safety measures like time-to-collision (TTC) (\cite{minderhoud2001extended}) and modified time-to-collision (\cite{ozbay2008derivation}), deceleration to avoid crash (DRAC) (\cite{fu2021comparison}), post-encroachment time (PET) (\cite{cooper1984experience}) and time-to-lane-crossing (TLC) (\cite{mammar2004time}). These measures use straightforward physical formulas to evaluate a vehicle's proximity to potential safety-critical events, such as collisions or lane departures. Another category focuses on subjective factors related to driver behavior and cognition in response to the risk. These factors include steering angles, braking intensity, and physiological states such as eye movements and blood volume pulse (\cite{petit2021risk}). These subjective factors assess driving risks by quantifying the pressure experienced by drivers or the intensity of their reactions to perceived hazards.

Both objective and subjective factors play a crucial role in driving risk assessment, particularly in emerging vehicle safety applications. For instance, while real-time risk assessment models must assist autonomous vehicles and ADAS in achieving the goal of near-zero accidents, they should also understand passenger risk perception to enhance their sense of security and comfort. Additionally, these models should account for the risk perception of other road users to prevent behaviors that might make human drivers feel unsafe, thereby contributing to the overall safety conditions of traffic.

However, both objective and subjective approaches for driving safety assessment have their limitations. First, collision risk models based on surrogate safety measures are often tailored to specific scenarios and lack generalizability across different use cases. Moreover, their effectiveness in assessing collision risks can vary across different facilities due to the heterogeneity of crash characteristics among them. On the other hand, subjective metrics, such as braking and steering behavior, are challenging to interpret without well-defined thresholds or appropriate aggregation methods (\cite{wang2021review}). Additionally, data on drivers' physiological states require specialized instruments for collection and can vary significantly between individuals (\cite{bao2020personalized}).

Recent studies have proposed the safety potential field method for driving risk assessment. This approach offers the advantage of providing a spatially continuous representation of safety risks across two-dimensional space, integrating both subjective and objective aspects of driving risks. It quantifies the spatial distribution of driving risk on the road using principles inspired by physical foundations, such as how a particle in an electric field experiences potential energy or a repulsive force relative to another charged object (\cite{wang2015driving}).  

However, existing safety potential field methods still face challenges and limitations that undermine their ability to accurately quantify highway driving safety risks:
\begin{itemize} 
\item First, existing safety potential field models utilize arbitrary physical functions that might inadequately represent the spatial distribution of risk from a driver’s perspective. As a result, these models lack a human-like understanding of risk and are limited in their ability to interpret driver behaviors. Even with extensive calibration, they struggle to realistically capture driver behavior, particularly in lateral maneuver scenarios.
\item Second, the scarcity of collision and near-miss data raises concerns about the effectiveness of safety potential field calibration. Variations in crash patterns and characteristics across different times and regions further undermine the generalizability of models calibrated using this type of event data. Moreover, the even rarer occurrence of lateral risk-related events presents further significant challenges for calibrating the lateral components of these models, raising doubts about their ability to accurately model lateral risks. 
\end{itemize}

To address the issues identified above, we propose a novel composite safety potential field (C-SPF) framework for comprehensive highway driving risk assessment.
C-SPF uses a composite structure, overlaying two separate safety potential field models, which evaluate driving risks from both subjective and objective perspectives. The C-SPF subjective field (S-field) is designed to quantify proximity risks beyond drivers' proximity tolerance, while the objective field (O-field) assesses imminent collision probability based on vehicles' relative motions. 

A key advantage of the C-SPF compared to existing models is that it does not rely on rare extreme event data for calibration. Instead, it uses vehicle spacing data to calibrate drivers' two-dimensional spatial proximity tolerance for the S-field, which is abundant in public naturalistic driving datasets and reflects drivers' proximity-based risk perception. On the other hand, the O-field in C-SPF uses straightforward motion functions to derive physical vehicle collision probabilities instead of using a psychological model requiring complex calibration, allowing for a direct evaluation of collision risks.

The proposed C-SPF combines the S-field and O-field and provides a comprehensive framework for assessing driving safety by addressing both subjective and objective risk dimensions.  Such a composite structure for risk assessment enables the integration of both the potential physical collision and the human perception factors. This integration improves the model's relevance and application in real-world scenarios, where driver actions and collision risks are closely interconnected.

The proposed C-SPF is calibrated and validated using the Highway Drone Dataset of Naturalistic Trajectories (highD) to ensure real-world applicability. Extensive analysis is performed to ensure that the C-SPF effectively captures risky scenarios and the corresponding driver behaviors. Case studies featuring multiple scenarios involving both longitudinal and lateral vehicle interactions are presented to verify the model's generalizability across various situations. The C-SPF is compared against existing safety measures, including a literature-based risk field (\cite{wang2016driving}). The results demonstrate that the C-SPF framework and its associated risk indicators effectively capture both longitudinal and lateral risks that triggered drivers' safety maneuvers. Additionally, the model outperforms existing safety measures in interpreting driving behaviors across various scenarios, such as car-following, lane-changing, and lateral adjustments in the lane. 

The remainder of the paper is organized as follows. Section \ref{sec:literature} reviewed related literature. Section \ref{sec:methodology} presents the methodology to structure the C-SPF. Section \ref{sec:experiment} describes the experiment settings and analyzes the results. Section \ref{sec:discussion} presents the discussion. Section \ref{sec:conclusion} presents the conclusion.

\section{Literature review}\label{sec:literature}
\subsection{Driving risk assessment}
Driving risk assessments utilize metrics or algorithms to identify potential hazards and evaluate their severity or proximity. These metrics or algorithms may incorporate objective factors, subjective factors, or both.

One commonly used objective metric is the Surrogate Safety Measure (SSM). These measures evaluate collision risk by analyzing traffic conflicts—situations that may not result in crashes but could lead to actual collisions if the involved road users do not alter their movements (\cite{arun2021systematic}). Some SSMs also assess additional risks, such as lane departure (\cite{mammar2004time}). These metrics have been widely used to provide real-time risk assessment inputs to various technologies to improve vehicle and traffic safety, such as adaptive signal control, collision-warning, advanced driving assistance system (ADAS), and autonomous driving. 

However, the literature identifies several limitations of existing SSMs for risk assessment. First, the correlation between SSMs and collisions is highly context-dependent. However, the correlation of these measures to actual accident statistics can vary widely across different facilities or even among segments of the same highway. This variability arises from the heterogeneity of crash characteristics, including types and occurrence patterns, across different types of facilities and regions. Existing safety assessment models often rely on thresholds that are either arbitrarily determined or calibrated using location-specific accident data, which undermines their generalizability across different facilities or road segments (\cite{vasudevan2022algorithms}). Second, conventional SSMs are designed for specific types of conflicts, and risk assessment models based on a single metric often lack applicability across different use cases. For instance, DRAC is specifically tailored for car-following scenarios, where lateral movements are ignored. Conversely, PET is suited for lateral interactions, such as turning and merging, making it more applicable to intersections and ramp areas (\cite{wang2021review}). 

Some models also assess driving risk based on subjective factors, evaluating drivers' perception of risks through behaviors, physiological states, or biometric data. \cite{phase2006100}, \cite{fazeen2012safe}, and \cite{stipancic2018vehicle} evaluated driving safety through vehicles' acceleration and deceleration behaviors, including instances of fast acceleration and hard braking. \cite{shangguan2020investigating} used drivers' braking reaction time to identify risky trajectories and quantify the impact of foggy weather on highway driving risk. \cite{petit2021risk} used the maximum steering angle to assess driving risk levels and utilized a learning-based method to study the influence of various vehicle and driver factors on safety. 

However, these subjective risk assessment methods also have drawbacks. First, driver behavior data, such as maximum deceleration rate, braking time, or steering angle, is typically used to evaluate driving risks over complete trajectories. Since these metrics are consequences of risk perception, they cannot proactively estimate risks, compromising their feasibility for real-time applications. Second, similar to the objective measures, most subjective measures derived from driver behavior are context-dependent. They rely on pre-determined thresholds to identify risk levels, but the optimal values of these thresholds can vary depending on driving styles and environments.

\subsection{Safety potential field}
The safety potential field method quantifies driving risks originated from artificial potential field theory, which is a modeling approach to quantify the risks associated with spatial proximity between objects. This theory was first applied to reactive navigation and path planning for robots by \cite{khatib1986real}. It utilized the repulsive force generated by obstacles within a virtually generated potential field to model collision avoidance behaviors. In the last decade, the potential field theory has been applied to a wide range of applications in various topics of transportation research not only in road safety and vehicle dynamics modeling but also in other fields such as pedestrian modeling \cite{helbing1995social} \cite{chraibi2011force} \cite{wang2016modified} \cite{yu2022pedestrian} \cite{tan2023rcp} and maritime navigation \cite{gan2022ship} \cite{shi2007harmonic}. 

In the realm of driving risk assessment, various safety potential field models have been proposed, employing distinct field functions grounded in physical principles to capture the spatial distribution of driving risk around vehicles. The existing safety potential field models can be classified as using either a subjective or objective approach, depending on their modeling approach. Most safety potential field models adopt a subjective psychological field with a physical foundation to represent drivers' risk perception. These models quantify how physical factors, such as proximity to collisions or distance to obstacles, influence subjectively perceived driving risks. For instance, \cite{wang2016driving} employed the principles of the Doppler effect and potential energy fields to define risk evolution based on changes in vehicle kinematics and various driver and road factors. Similarly, \cite{li2020dynamic}incorporated acceleration to refine the field's shape and modeled drivers' proximity risk perception as a repulsive force. Additionally, \cite{arun2023physics} introduced a psychological risk force combining the risk of proximity to an impending collision and the expected severity of a collision. 

On the other hand, \cite{mullakkal2020probabilistic} proposed an objective approach using a collision probabilistic field considering solely physical kinematic factors, such as vehicles' feasible deceleration range, to assess vehicles' driving risk by objectively predicting their collision probability. \cite{wu2019modified} proposed a driving safety field to estimate collision risk between vehicles and pedestrians at unsignalized road sections by incorporating the randomness in pedestrian motion predictions. \cite{song2024subjective} proposed a driving risk prediction model that integrates a cognitive field representing the human driver's perspective with an objective anisotropic risk field to quantify the risk posed by nearby risk entities to achieve human-like risk assessment. 

Existing studies employ various methods to calibrate their safety potential field models. \cite{wang2016driving} discussed the use of traffic accident statistics for model calibration. \cite{arun2023physics} utilized crash characteristic data from intersections within a city to fit their model. \cite{li2020dynamic} and \cite{tan2021risk} calibrated their car-following models using deceleration behaviors derived from trajectory datasets. Additionally, \cite{song2024subjective} employed a learning-based approach to fit their model using risk-level data manually labeled for lane-changing videos.
 
Existing safety potential field models have been applied to various risk assessment tasks, leveraging their unique features and strengths. \cite{wang2016driving} and \cite{mullakkal2020probabilistic} investigated risks related to car-following and cut-in maneuvers using test vehicles and simulations. \cite{li2020risk} and \cite{wang2024toward} used safety potential fields to assess collision risks associated with lane blockages and work zones through simulations. \cite{chen2022modeling} and \cite{zhang2024real} employed field-based approaches to assess rear-end collision risks in tunnel areas, while \cite{song2022dynamic} evaluated the risk of secondary rear-end collisions. \cite{song2024subjective} focused on predicting subjectively labeled risk levels associated with lane-changing maneuvers.


\section{Composite Safety Potential Field}\label{sec:methodology}

\subsection{Subjective field}
The first component of C-SPF, the Subjective Field (S-field), quantifies drivers' perceived risk associated with spatial proximity to nearby risk entities. 
The S-field assumes that each highway vehicle proactively maintains a safety space around itself such that any other vehicle or object perceived to intrude upon this space poses a proximity risk and will compel the driver to adjust the spacing. The closer proximity to the intruding risk entity results in stronger risk. 

To model the naturalistic drivers' behavior to maintain such a proactive safety space and estimate the risk caused by the intrusion, we adopted the concept of probabilistic spacing proposed by \cite{jiao2023inferring}. This approach defines proximity resistance, equivalent to the proximity risk in the proposed S-field, as one minus the proximity tolerance, where proximity tolerance is defined as the likelihood of the presence of the observed surrounding vehicles. The S-field risk intensifies with diminishing relative spacing, with a higher risk value reflecting greater pressure or discomfort perceived by the drivers, encouraging the driver to adjust the relative distance. While \cite{jiao2023inferring} focuses solely on inter-vehicle spacings, the proposed S-field extends its analysis to include risk entities such as vehicles, lane markers, and road boundaries.

\subsubsection{Vehicle proximity risk}\label{sec:ggd}

The generalized Gaussian distribution (GGD) was identified by \cite{jiao2023inferring} as an effective method for inferring proximity resistance based on drivers' probabilistic spacing. The original one-dimensional GGD probability density function writes (\cite{novey2009complex}),
\begin{equation}
\label{ggd}
f(x|\mu, \gamma, \beta)=\frac{\beta}{2\gamma \Gamma(1/\beta)} \exp\left[ - \left( \frac{x-\mu}{\gamma} \right)^\beta \right] 
\end{equation}
where the parameters $\mu$, $\gamma$, and $\beta$ are the mean, scale factor, and shape factor of GGD. $\Gamma(1/\beta)$ represents a Gamma distribution with shape parameter $1/\beta$. 

\begin{figure}
\centering
\includegraphics[width=.65\textwidth]{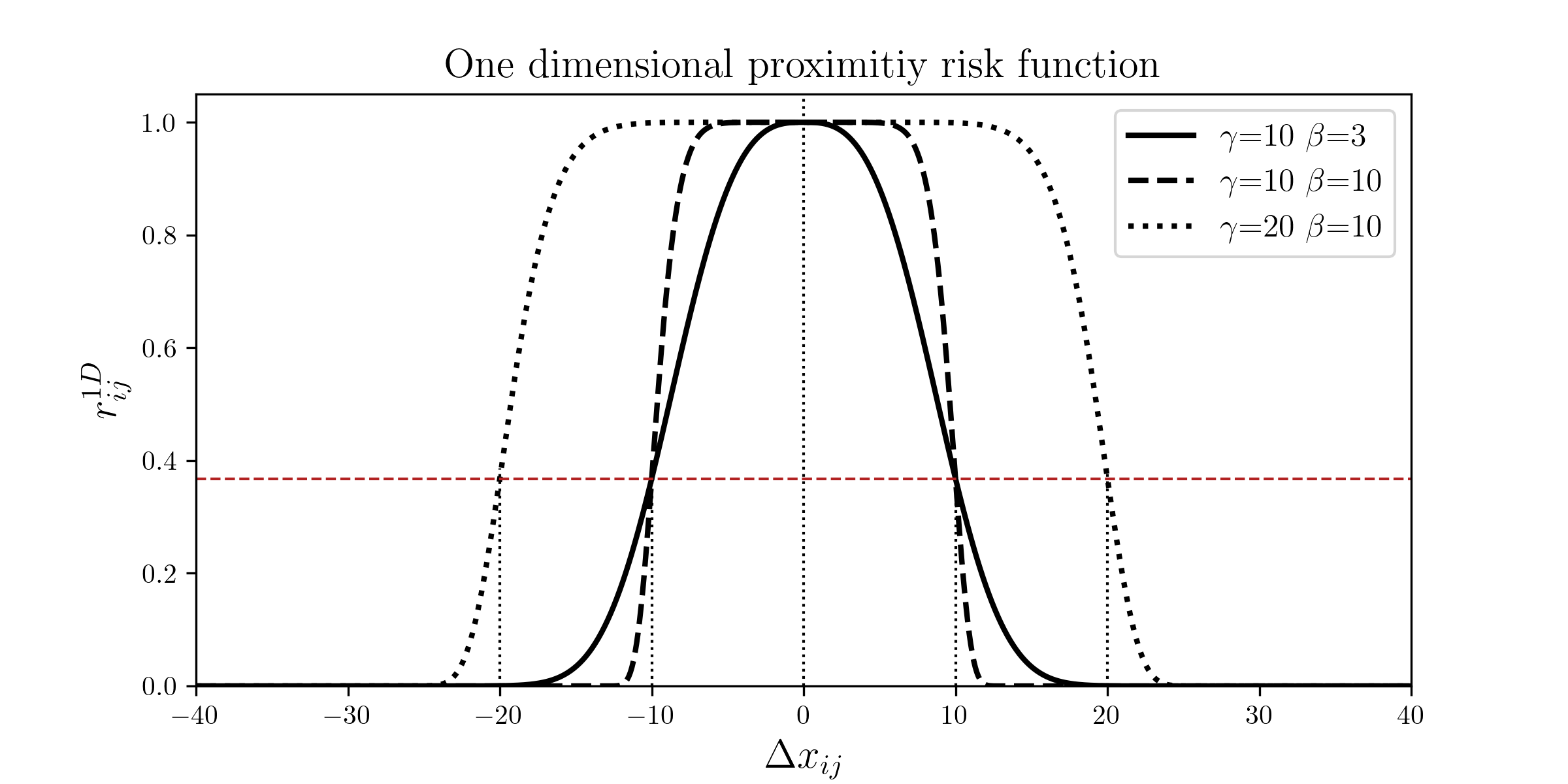}
\caption{One-dimensional proximity risk distribution based on GGD}
\label{fig:ggd-1d}
\end{figure}

To estimate the proximity risk perceived by drivers, equation (\ref{ggd}) is scaled between 0 and 1. The one-dimensional S-field risk perceived by the ego vehicle $j$ due to its spacing with a nearby vehicle $i$ is defined as:  
\begin{equation}
\label{ggd_risk_1d}
r_{s,ij}^{1D}= \exp\left( - \left| \frac{\Delta x_{ij}}{\gamma} \right|^\beta \right)
\end{equation}
where $r_{s,ij}^{1D}$ represents the one-dimensional proximity risk perceived by the ego vehicle $j$ due to the spacing with another vehicle $i$ nearby. $\Delta x_{ij}$ represents the longitudinal distance between the two vehicles. Figure \ref{fig:ggd-1d} illustrates the shapes of the one-dimensional proximity risk represented by GGD functions with different parameters. The red dashed line in Figure \ref{fig:ggd-1d} indicates the risk value at a spacing of $\Delta x_{ij}=\gamma$. The scale factor $\gamma > 0$ defines the critical safety distance beyond which drivers begin to perceive escalating risks, while $\beta \geq 2$ governs the rate at which the proximity risk escalates. 
The two-dimensional S-field function is based on the extension of Eq. (\ref{ggd_risk_1d}),
\begin{equation}
\label{ggd_risk_2d}
r_{s,ij}^{v}= \exp\left( - \left| \frac{\Delta x_{ij}}{\gamma_x} \right|^{\beta_x}  - \left| \frac{\Delta y_{ij}}{\gamma_y} \right|^{\beta_y}  \right)
\end{equation}
where $r_{s,ij}^{v}$ denotes the two-dimensional proximity risk perceived by the ego vehicle $j$ due to perceiving the presence of nearby vehicle $i$. $\Delta x_{ij}$ represents the longitudinal relative distance between the two vehicles along the direction of motion of the ego vehicle $j$, while $\Delta y_{ij}$ represents the lateral relative distance between the two vehicles measured perpendicular to the ego vehicle's direction of motion. Both $\Delta x_{ij}$ and $\Delta y_{ij}$ are measured around between the closest points between the two vehicles. $\gamma_x > 0$ and $\gamma_y > 0$ are scale factors along the longitudinal and lateral directions, respectively, while $\beta_x \geq 2$ and $\beta_y \geq 2$ are the corresponding shape factors. 
\begin{figure}
\label{fig:dx dy}
\end{figure}

\subsubsection{Lane marker and boundary proximity risk}
The ego vehicle $j$'s S-field risk relative to the near lane marker $a$ is defined as a function of their lateral distance in between:
\begin{equation}
\label{ggd_risk_lane}
r_{s,aj}^{l} = \exp\left( - \left| \frac{\Delta y_{aj}}{\gamma_l}  \right|^{\beta_l}  \right)
\end{equation}
where $\gamma_l>0$ and $\beta_l \geq 2$ denote the scale factor and shape factor, respectively. $\Delta y_{aj}$ denotes the lateral distance between lane marker $a$ and the center of the ego vehicle $j$. 

Similarly, the ego vehicle $j$'s S-field risk relative to the near road boundary $k$ is defined as follows:
\begin{equation}
\label{ggd_risk_boundary}
r_{s,kj}^{b} = \exp\left( - \left| \frac{\Delta y_{kj}}{\gamma_b}  \right|^{\beta_b}  \right)
\end{equation}
where $\gamma_b>0$ and $\beta_b \geq 2$ denote the scale factor and shape factor, respectively. $\Delta y_{kj}$ denotes the lateral distance between road boundary $k$ and the center of the ego vehicle $j$. 
\begin{figure}
\label{fig:dy}
\end{figure}

\subsubsection{Parameter inference}
\label{sec:parameter inference}
The parameters of the S-field in Eqs. (\ref{ggd_risk_2d}), (\ref{ggd_risk_lane}), and (\ref{ggd_risk_boundary}) should be derived from naturalistic driving data to achieve a human-like estimation of drivers' safety space across different vehicle velocities. Similar to the parameter inference method described in \cite{jiao2023inferring}, the S-field parameters were derived by estimating the density distribution of the accumulated presence of surrounding risk entities. This approach assumes that more frequent proximity in the distribution under similar conditions of vehicle kinematics is generally more tolerable to drivers and, therefore, poses a lower S-field risk. This approach assumes that more frequent proximities in the distribution, under the same conditions, are generally more tolerable to drivers and thus pose a lower S-field risk. Conversely, less frequent proximities in the distribution correspond to higher S-field risks. Therefore, the S-field, as defined in Eqs. (\ref{ggd_risk_2d}), (\ref{ggd_risk_lane}), and (\ref{ggd_risk_boundary}), should be calibrated to effectively differentiate between frequent and infrequent spacings observed across a large population of drivers.

The likelihood of the observed spacing, based on the definition of proximity risk, can be expressed as one minus the corresponding S-field risk value. By aggregating spacing data over extended time frames and sample vehicles, the joint likelihood of the accumulated spacing samples is defined as:
\begin{equation}
\label{eq:joint likelihood}
L = \prod_j^{n} \left[ \prod_{i=1}^{N_j^v}\left(1-r_{s,ij}^{v}\right)\prod_{a=1}^{N_j^l}\left(1-r_{s,aj}^{l}\right)\prod_{k=1}^{N_j^b}\left(1-r_{s,kj}^{b}\right) \right]
\end{equation}
where $N_j^v$, $N_j^l$, and $N_j^b$ represent the number of nearby vehicles perceived, lane markers, and road boundaries, respectively. One minus the S-field risk, such as $1-r_{s,ij}^{v}$, $1-r_{s,aj}^{l}$, and $1-r_{s,kj}^{b}$, represents the likelihood of the current spacing relative to the corresponding risk entity. $n$ denotes the total number of accumulated vehicle frames.

The joint log-likelihood can be written as,
\begin{equation}
\label{eq:joint loglikelihood}
\ln(L) = \sum_j^{n} \left[ \sum_{i=1}^{N_j^v} \ln\left(1-r_{s,ij}^{v}\right) + \sum_{a=1}^{N_j^l} \ln\left(1-r_{s,aj}^{l}\right) + \sum_{k=1}^{N_j^b} \ln\left(1-r_{s,kj}^{b}\right) \right]
\end{equation}

According to \cite{jiao2023inferring}, the optimal $\beta$ should maximize the joint likelihood $L$ to capture the sparsity of infrequent proximity. Meanwhile, the optimal $\gamma$ should identify typical drivers' critical safety space, representing the distance with which other vehicles and objects rarely intrude. Specifically, $\gamma$ should identify the position where the likelihood escalated transitions from a gradual to a rapid increase as the distance diminishes, corresponding to the point where the second-order derivative of likelihood $L$ with respect to $\gamma$ is minimized.

Thus, inferring S-field parameters involves iteratively solving the following problems until all parameters converge or the maximum number of iterations is reached:
\begin{equation}
\label{estimate_beta}
\hat{\beta} \leftarrow \underset{\beta}{\mathrm{argmax}} \ln(L) \quad \textrm{s.t.} \:\: \beta \geq 2 \quad \text{for $\beta$ =  $\beta_x$, $\beta_y$, $\beta_l$, $\beta_b$ }
\end{equation}
and
\begin{equation}
\label{estimate_gamma}
\hat{\gamma} \leftarrow \underset{\gamma}{\mathrm{argmin}} \frac{\partial}{\partial \gamma}\left(\frac{\partial\ln(L)}{\partial\gamma}\right) \quad \textrm{s.t.} \:\: \gamma > 0  \quad \text{for $\gamma$ = $\gamma_x$, $\gamma_y$, $\gamma_l$, $\gamma_b$ }
\end{equation}

\subsubsection{Influencing factors of field shape}
The absolute velocity is identified as the key factor shaping the proposed S-field. The influence of vehicle velocity on drivers' subjective safety space aligns with established research demonstrating its impact on both longitudinal and lateral spacing in highway environments. For instance, vehicles traveling at higher speeds tend to maintain greater space headways with preceding vehicles during car-following, even in the absence of speed differences (\cite{lazar2016review}). This behavior reflects the longitudinal aspect of the safety space drivers proactively maintain. Additionally, studies have shown that absolute velocity correlates with inter-vehicle lateral distance  (\cite{delpiano2015characteristics}) and vehicles' lateral positioning within the lane (\cite{dai2022research}), illustrating the lateral dimension of this subjective safety space.

Other factors, such as speed difference, vehicle class, road and environmental conditions (e.g., lighting, road curvature, and pavement quality), and driver characteristics (e.g., driving skills and age), are not addressed in the S-field. First, although speed difference is intuitively related to drivers' risk perception, it is already incorporated into the second component of C-SPF, the O-field. Therefore, it is not selected for the S-field to avoid redundancy. Second, incorporating multiple risk factors complicates the calibration process. The methodology in Section \ref{sec:parameter inference} estimates $\gamma$ and $\beta$ directly, rather than their relationship with risk factors. Establishing this relationship requires discretizing the risk factors and segmenting the dataset into smaller divisions based on the factor combinations. Calibration is then performed separately for each division. However, increasing the dimensionality of factors significantly reduces the data available for each division, which can substantially compromise the reliability of the calibration results.

\subsubsection{Aggregated subjective risk}
The aggregated S-field risk experienced by the ego vehicle $j$ in a highway environment surrounded by multiple vehicles is defined as one minus the joint likelihood of the ego vehicle's spacing relative to all surrounding risk entities, such that: 
\begin{equation}
\label{eq:agg s-risk}
r_{s,j} =  1 - \prod_{i=1}^{N_j^v}(1-r_{s,ij}^{v})\prod_{a=1}^{N_j^l}(1-\kappa_l r_{s,aj}^{l})\prod_{k=1}^{N_j^b}(1-\kappa_b r_{s,kj}^{b})
\end{equation}
where $\kappa_l$ and $\kappa_b$ are coefficients between 0 and 
1 to balance the relative risk contribution of lane markers and boundaries compared to vehicles, since crossing the lane marker usually does not cause a collision, nor does crossing a road boundary if it is not a physical barrier. $r_{s,j}$ rises as the number of vehicles nearby increases, or as the risk associated with any risk entity escalates.

\subsection{Objective field}

The second component of C-SPF, the objective field (O-field), measured the risk of collision by estimating the probability that two risk entities would collide during the process of relative movements.
\subsubsection{Generalized collision risk}
Intuitively, two vehicles suffer elevated collision probability in a two-dimensional space when: (1) there is a potential collision point in their predicted future trajectory, (2) there is insufficient time for drivers' maneuver to avoid the collision. Based on this rationale, the O-field is defined as follows,
\begin{equation}
\label{eq:o-risk}
r_{o,ij} = P_{ij} \cdot T_{ij}
\end{equation}
where $r_{o,ij}$ denotes the objective collision risk estimated by O-field between vehicle $i$ and $j$, which will be referred to as O-field risk in the remainder of the paper. $P$ is a spatial proximity factor that predicts whether the nearest future proximity between two vehicles could potentially lead to a collision. $T$ is a temporal proximity factor that quantifies the time proximity to that potential collision. For simplicity and better generalization, the O-field measured the collision risk at discrete times and estimated $P$ and $T$ from vehicles' current two-dimensional distances and relative kinematics. 

To estimate $P$ at a specific time point, the O-field predicts the positions where the relative distance between two risk vehicles will be minimized in the near future and determines whether this minimal future distance would result in a collision. Intuitively, a collision will be highly likely if that minimal future distance between the two vehicles falls below a predefined threshold. Therefore, the definition of $P$ is given as follow,
\begin{equation}
\label{eq:P}
P_{ij} = \exp \left[ -\left(\frac{\hat{d}_{m,ij}}{d^*} \right)^{\beta_p} \right]
\end{equation}
where $\hat{d}_{m,ij}$ denotes the predicted minimum future distance between risk vehicles $i$ and $j$ from the risk estimation point into the near future during the process they approach each other. $d^*$ is a predefined distance threshold to check the occurrence of collision. $\beta_p$ is a shape factor. 

Let $\hat{d}_{ij}(t)$ represent the predicted future distance between two vehicles at time $t$, measured from the risk estimation point where $t=0$. If the two vehicles exhibit a tendency to approach each other at the risk estimation point, there will be time derivative $\dot{\hat{d}}_{ij} < 0$ at $t=0$. Then the predicted future minimum distance $\hat{d}_{m,ij}$ can be defined as,
\begin{equation}
\label{eq:d_m}
\hat{d}_{m,ij} = \begin{cases}
0 & \text {if $\hat{d}_{ij}(0) = 0$}\\
\hat{d}_{ij}(\hat{t}_{m,ij})  & \text{if $\hat{d}_{ij}(0) > 0$ and $\dot{\hat{d}}_{ij}(0) < 0$ } \\
\infty & \text{else}
\end{cases}
\end{equation}
where $\hat{t}_{m,ij}$ is defined as the time frame when the distance between the vehicle stops decreasing, which can be expressed as follows,
\begin{equation}
\label{eq:t_m}
\hat{t}_{m,ij} =
\begin{cases}
0 & \text {if $\hat{d}_{ij}(0) = 0$} \\
\max \ t \quad \textrm{s.t.} \:\:  \dot{\hat{d}}_{ij}(t')<0, \:\: \forall \text{ $0\leq t'< t$}. & \text{if $\hat{d}_{ij}(0) > 0$ and $\dot{\hat{d}}_{ij}(0) < 0$ }  \\ 
\infty & \text {otherwise}
\end{cases}
\end{equation}

Similar to how $P_{ij}$ is estimated in Eq. (\ref{eq:P}), we can define the time proximity $T_{ij}$ to the potential collision point as a function of $\hat{t}_{m,ij}$, which writes,
\begin{equation}
\label{T}
T_{ij} = \exp \left[ -\left(\frac{\hat{t}_{m,ij}}{t^*} \right)^{\beta_t} \right]
\end{equation}
where $t^*$ is a scaling factor and ${\beta_t}$ is a shape factor. The value of $T_{ij}$ ranges from 0 to 1, with a lower $\hat{t}_{m,ij}$ producing a higher time intensity to the potential collision point and, thus, a greater collision risk. $T_{ij}$ equals 1 if a collision has already happened, indicated by $\hat{d}_{ij}(0)=0$, and equals 0 if two vehicles are moving away from each other since the risk estimation point.

\subsubsection{Simplified objective risk}

For simplicity, vehicles are assumed to maintain constant velocities and directions of movement in a short period. Under this assumption, the relationship between the predicted future distance $\hat{d}_{ij}(t)$ between two risk vehicles and their current two-dimensional distances and instantaneous speed differences can be expressed as, 
\begin{equation}
\label{d}
\hat{d}_{ij}(t) =  \left| \boldsymbol{D}_{ij} + \boldsymbol{V}_{ij} \cdot t \right|, \quad t \geq 0
\end{equation}
where $\boldsymbol{D}_{ij}$ is the distance vector between vehicle $i$ and $j$ at the point of risk estimation, 
$
\boldsymbol{D}_{ij} = \begin{bmatrix}
           x_{i} - x_{j} \\
           y_{i} - y_{j} \\
         \end{bmatrix}
         = \begin{bmatrix}
           \Delta x_{ij} \\
           \Delta y_{ij} \\
         \end{bmatrix}
$. 
$x_{i}$ and $y_{i}$ are the x and y coordinate of vehicle $i$ on the road. 
$\boldsymbol{V}_{ij}$ is the speed difference vector between vehicle $i$ and $j$ at the point of risk estimation, 
$
\boldsymbol{V}_{ij} = \begin{bmatrix} 
           v_{x,i} - v_{x,j} \\
           v_{y,i} - v_{y,j} \\ 
         \end{bmatrix}
        = \begin{bmatrix} 
           \Delta v_{x,ij} \\
           \Delta v_{y,ij} \\ 
         \end{bmatrix}
$. $v_{x,i}$ and $v_{y,i}$ are the speed factor of vehicle $i$ along x and y axis, respectively. 



Substitute Eq. (\ref{d}) into Eq. (\ref{eq:t_m}) and solve under the given conditions. The computation of $\hat{t}_{m,ij}$ is then adjusted as follows,
\begin{equation}
\label{tm}
\hat{t}_{m,ij} = \begin{cases} 
0 & \text{if $\left| \boldsymbol{D}_{ij} \right| = 0 $} \\
 -\frac{\boldsymbol{D}_{ij}^{T}\boldsymbol{V}_{ij}}{\boldsymbol{V}_{ij}^{T}\boldsymbol{V}_{ij}}  =  -\frac{\Delta x_{ij}\Delta v_{x,ij}+\Delta y_{ij}\Delta v_{y,ij}}{\Delta v_{x,ij}^2+\Delta v_{y,ij}^2}   & \text{if  $\boldsymbol{D}_{ij}^{T}\boldsymbol{V}_{ij} < 0$ and $\left| \boldsymbol{D}_{ij} \right| > 0 $ } \\

\infty    & \text{otherwise}
\end{cases}
\end{equation}

Combining Eq. (\ref{tm}) and Eq. (\ref{eq:d_m}) will give the adjusted expression of $\hat{d}_{m,ij}$ under the assumption of constant velocities:
\begin{equation}
\label{eq:dm_linear}
\hat{d}_{m,ij} = \begin{cases}
0 & \text{if $\left| \boldsymbol{D}_{ij} \right| = 0 $} \\
\frac{\left| 
\boldsymbol{D}_{ij}^{T}\boldsymbol{A}\boldsymbol{V}_{ij}  \right|}{\left| \boldsymbol{V}_{ij} \right|} = \left| \boldsymbol{D}_{ij}^{T}\boldsymbol{A} \hat{V}_{ij}  \right|  = \sqrt{ \frac{\left( \Delta y_{ij}\Delta v_{x,ij}-\Delta x_{ij}\Delta v_{y,ij} \right)^2}{\Delta v_{x,ij}^2+\Delta v_{y,ij}^2} } & \text{if $\boldsymbol{D}_{ij}^{T}\boldsymbol{V}_{ij} < 0$ and $\left| {\boldsymbol{D}}_{ij} \right| > 0 $ } \\
\infty & \text{otherwise}
\end{cases}
\end{equation}
where $\boldsymbol{A}$ is a transformation matrix which writes,
$
\label{A}
A = \begin{bmatrix}
          0 & 1  \\
          -1 & 0  \\
         \end{bmatrix}
$. 
$\hat{\boldsymbol{V}}_{ij}$ denotes the speed difference $\boldsymbol{V}_{ij}$ vector normalized by its length, such that $\hat{\boldsymbol{V}}_{ij} = \boldsymbol{V}_{ij} / |\boldsymbol{V}_{ij}|$. According to Eq.(\ref{eq:dm_linear}), the value of $\hat{d}_{m,ij}$ is irrelevant to the magnitude of the vehicles' speed difference under the assumption of constant velocities. Instead, it depends solely on their current distances and the direction of their relative speed.

Therefore, based on equation (\ref{eq:o-risk}), the O-field that quantifies vehicles' collision risk at any discrete time point, assuming constant speed differences, can be expressed as:
\begin{equation}
\label{eq:o-field-1}
r_{o,ij}=
\begin{cases}
1 & \text{if $\left| \boldsymbol{D}_{ij} \right| = 0 $} \\
\exp \left[ -\left(\frac{1}{d^*}\cdot \left| \boldsymbol{D}_{ij}^{T}\boldsymbol{A} \hat{V}_{ij}  \right| \right)^{\beta_p} \right] \cdot \exp \left[ -\left(\frac{1}{t^*}\cdot -\frac{\boldsymbol{D}_{ij}^{T}\boldsymbol{V}_{ij}}{\boldsymbol{V}_{ij}^{T}\boldsymbol{V}_{ij}}  \right)^{\beta_t} \right]  & \text{if $\boldsymbol{D}_{ij}^{T}\boldsymbol{V}_{ij} < 0$ and $\left| {\boldsymbol{D}}_{ij} \right| > 0 $ } \\
0 & \text{otherwise}    
\end{cases}
\end{equation}

\subsubsection{Aggregated objective risk}
In a scenario where the ego vehicle $j$ is surrounded by other multiple vehicles $i \in N_i$, the aggregated O-field risk posed on vehicle $j$ is defined as the probability of a collision occurring with any of the surrounding vehicles:
\begin{equation}
\label{eq:agg o-risk}
r_{o,j} = 1- \sum_{i \in N_j} (1-r_{o,ij})
\end{equation}

The aggregated O-field risk vehicle $j$ experiences will rise as the number of nearby vehicles increases, or as the collision risk with any of the surrounding vehicles escalates.





\section{Experiments}\label{sec:experiment}

\subsection{Data description}
The Highway Drone Dataset Naturalistic Trajectories (highD) (\cite{krajewski2018highd}) is used for the calibration and validation of C-SPF. The dataset is recorded on German highways using drones at six different locations, covering over 110,500 vehicles and each segment measuring 420 meters. The dataset includes detailed information for each vehicle extracted by computer vision algorithms, including position, velocity, acceleration, dimension, and vehicle sight distance. The data also include positions of lane markers and road boundaries. The data was collected between 8 AM and 5 PM during windless and sunny days allowing the influence of weather conditions on driving behaviors to be disregarded. The complete dataset of highD is used for parameter inference and analysis. 

\begin{figure}
        \centering
        \begin{subfigure}[H]{0.45\textwidth}
            \centering
            \includegraphics[width=\textwidth]{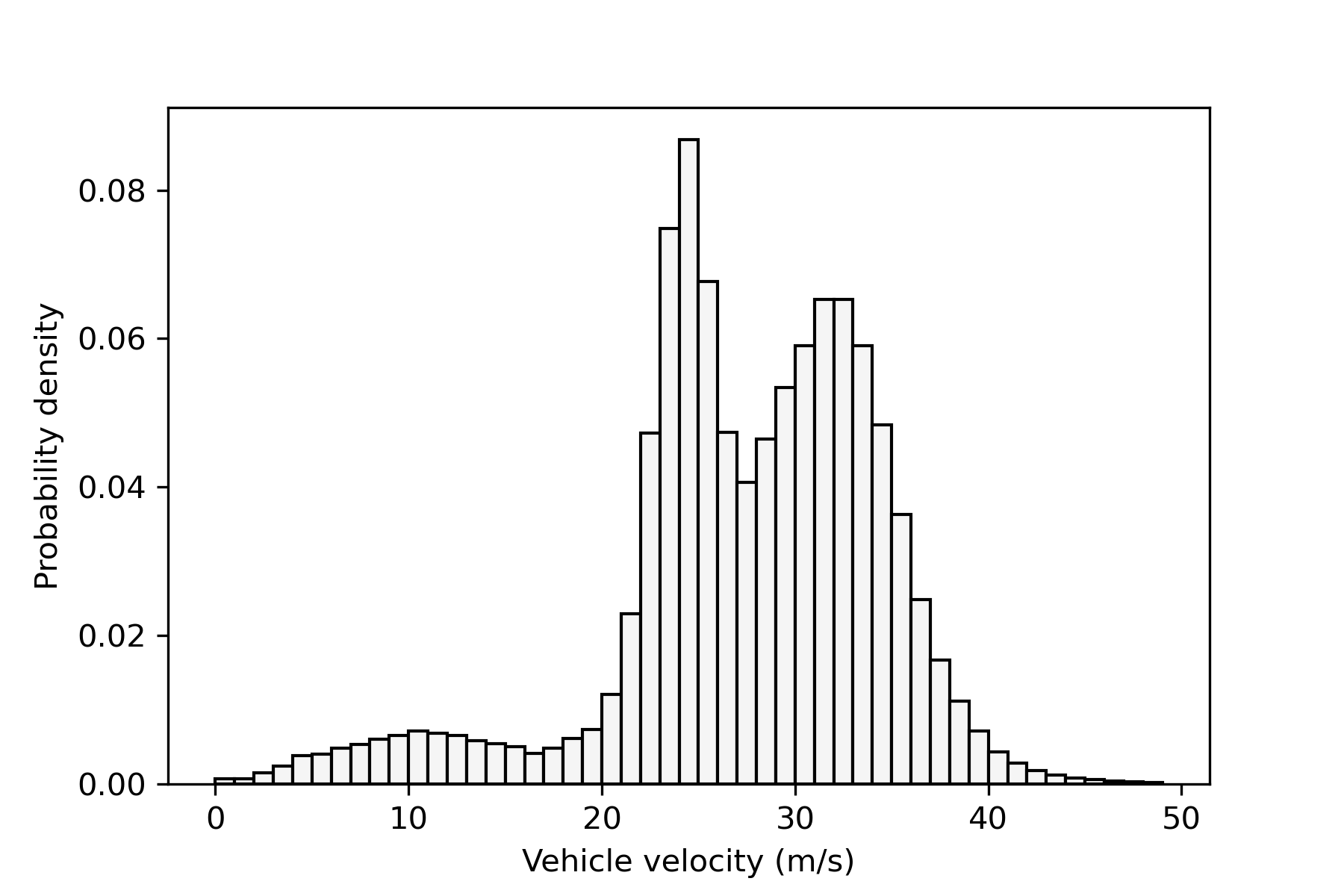}
            \caption[]%
            {{\small HighD vehicle-frame velocity distribution (m/s)}}    
            \label{fig:subj1}
        \end{subfigure}
        \begin{subfigure}[H]{0.45\textwidth}  
            \centering 
            \includegraphics[width=\textwidth]{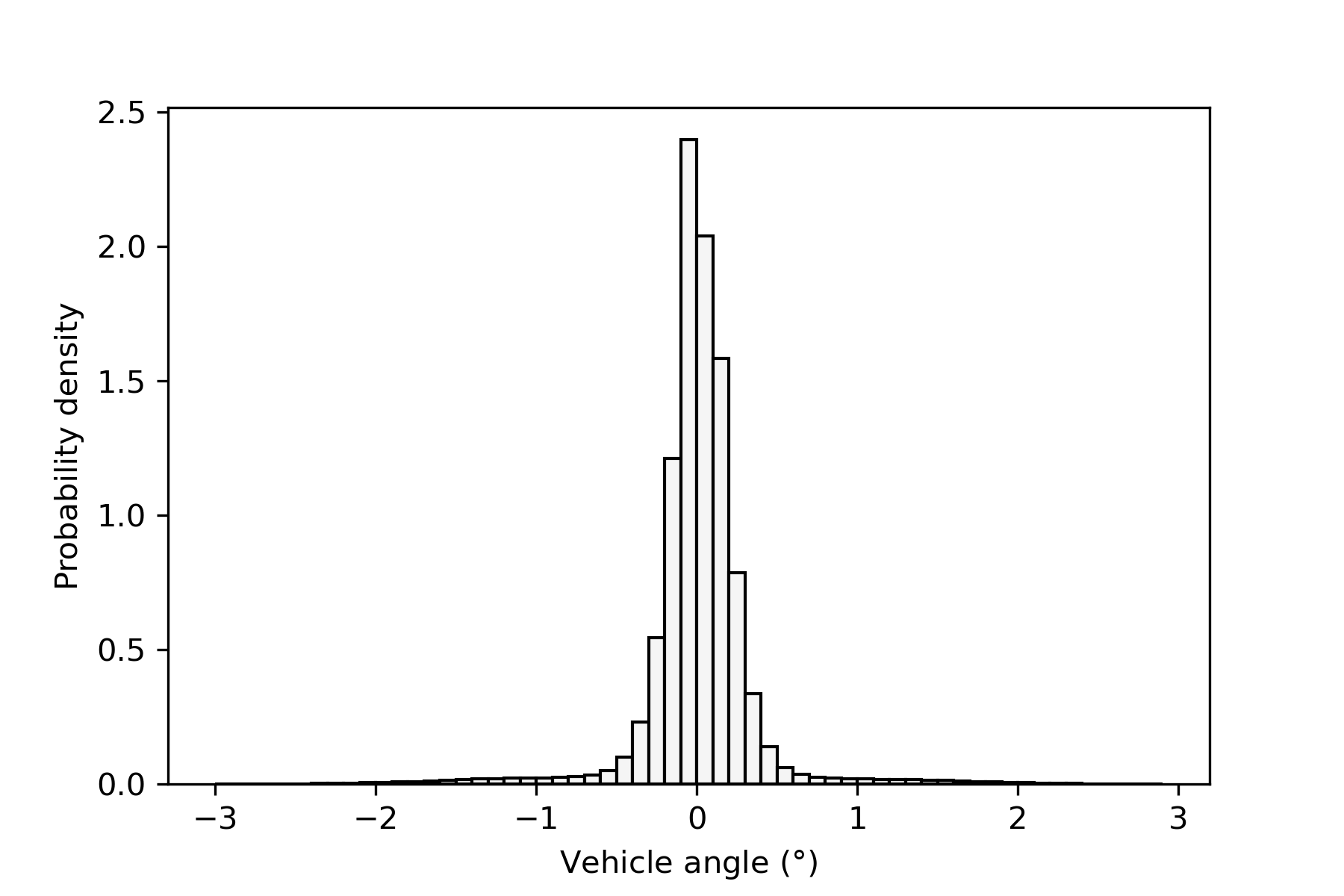}
            \caption[]%
            {{\small HighD vehicle-frame steering angle distribution ($\degree$)}}    
            \label{fig:subj2}
        \end{subfigure}
        \caption {Distribution of velocity and steering angle of highD vehicles}
        \label{fig:highd vel and ang}
    \end{figure}

Figure \ref{fig:highd vel and ang}(a) illustrates the distribution of vehicle velocity per vehicle frame in the highD dataset. The vehicle velocity ranges from 0 to 50 meters per second (m/s), with the majority of the frames concentrating between 20 and 40 m/s. The dataset also includes extensive trajectory data captured during congestion, with vehicle velocity nearing zero. Figure \ref{fig:highd vel and ang}(b) illustrates the distribution of clockwise angles of vehicles' moving direction, calculated by $\theta=180\degree /\pi \times\arcsin\left({v_y/\sqrt{v_x^2+v_y^2}}\right)$, where $v_x$ and $v_y$ denote the longitudinal and lateral speed factor of vehicle, respectively. Most vehicle angles are between -3$\degree$ and 3$\degree$.

\subsection{Model calibration}

\subsubsection{S-field parameters}

The S-field parameters are calibrated using the highD dataset. The entire dataset is segmented based on the closest integer number of the instantaneous velocity. For example, the velocity division of 20 m$/$s includes all trajectories with velocities ranging from 19.5 to 20.5 m$/$s. The parameter inference method described in Section \ref{sec:parameter inference} is applied to each individual division. To generalize across all vehicles, bootstrapping is used to create subsamples of different vehicles from each velocity-specific division (\cite{efron1994introduction}). In each velocity division, subsampling is repeated 20 times with 85\% of the vehicles in the division randomly selected with replacement for each iteration. The final parameters for each division are calculated as the means across all 20 iterations. 

\begin{figure}[h!]
\centering
	\includegraphics[width=.9\textwidth]{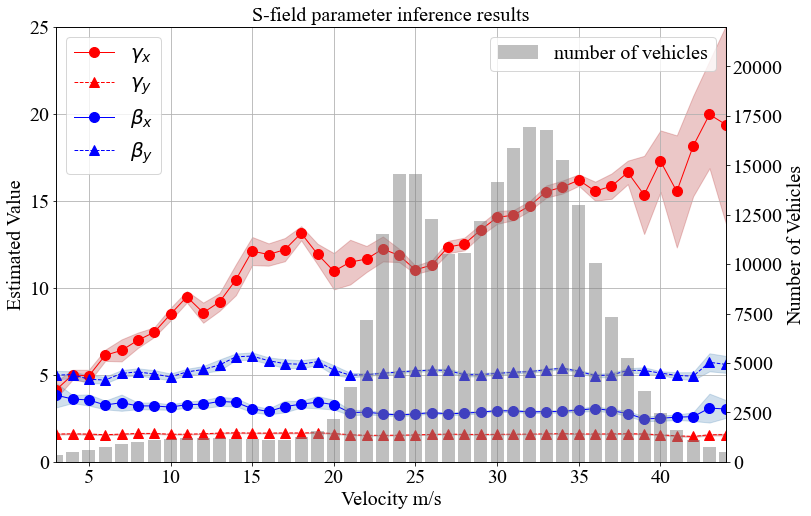}
	\caption{S-field vehicle proximity risk parameter inference results with bootstrapping sampling. The filling colors indicate the standard deviations obtained from bootstrapping.}
\label{fig:bootstrapping}
\end{figure}

Figure \ref{fig:bootstrapping} illustrates the parameter inference results. The data points represent the mean estimated values from 20 iterations of bootstrapping for each velocity division, where the data points represent mean estimated values from the 20 bootstrapping iterations and the shaded areas around the data points depict the standard deviation of the parameters derived from the bootstrapping process. The bar plots show the number of vehicles in each velocity division. However, the calibration method did not succeed for trajectory data points with velocities below 3 m$/$s or above 42 m$/$s due to insufficient data for calibration. Therefore, it is necessary to establish a continuous relationship between the S-field parameters and velocity using the calibrated parameters from regions where calibration is successful. The parameters for velocities outside these regions are then predicted based on the fitted relationship. 

According to Figure \ref{fig:bootstrapping}, $\gamma_x$ increases with velocity, reflecting the tendency of drivers to maintain greater spacing when traveling at higher speeds. The rate of increase in $\gamma_x$ with velocity is not always linear across different velocity ranges. 

To establish a continuous relationship between $\gamma_x$ and vehicle velocity, the following function is used for fitting the estimated parameters:
\begin{equation}
\label{eq:gamma_x}
\gamma_x = 5.1053\times10^{-4} \times v_j^3 -3.7051 \times10^{-2}\times v_j^2 + 1.0621\times v_j + 1.2925
\end{equation}

For $\beta_x$, the estimated mean values decrease with increasing velocity at a non-linear rate. This trend is opposite to $\gamma_x$. These findings suggest that while vehicles maintain a longer safety distance to the preceding vehicle at a higher velocity, the rate at which the proximity escalates with diminishing distance becomes lower, meaning they are more tolerant to vehicles intruding the safety space at $\Delta x=\gamma_X$. The following function is used to establish a continuous relationship between $\beta_x$ and vehicle velocity:
\begin{equation}
\label{eq:beta_x}
\beta_x = 2.2214\times10^{-5} \times v_j^3 -1.4834 \times10^{-3}\times v_j^2 + 9.6673\times10^{-3}\times v_j + 3.2589
\end{equation}

In the lateral dimension, the fitted $\gamma_y$ and $\beta_y$ show minimal variation with increasing velocity. Their mean values across all velocities, 1.4310 for $\gamma_y$ and 4.9956 for $\beta_y$, are chosen as the final estimated parameters. The value of $\beta_y$ is significantly higher than that of $\beta_x$, showing that drivers are more sensitive to the S-field risk as the spacing decreases per meter in the lateral dimension than in the longitudinal dimension. This can be explained by less interaction space in the lateral dimension than in the longitudinal dimension.

\begin{figure}
        \centering
        \begin{subfigure}[H]{0.30\textwidth}
            \centering
            \includegraphics[width=\textwidth]{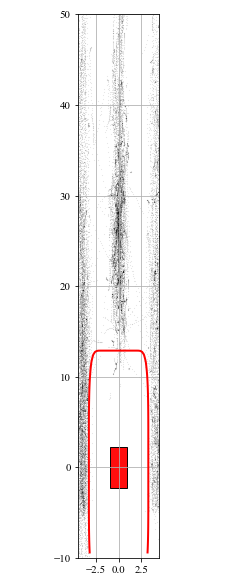}
            \caption[]%
            {{\small S-field at vehicle velocity 15 m$/$s}}    
            \label{fig:subj1}
        \end{subfigure}
        \begin{subfigure}[H]{0.30\textwidth}  
            \centering 
            \includegraphics[width=\textwidth]{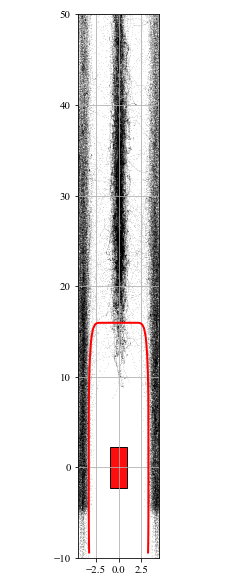}
            \caption[]%
            {{\small S-field at vehicle velocity 25 m$/$s}}    
            \label{fig:subj2}
        \end{subfigure}
        \begin{subfigure}[H]{0.30\textwidth}   
            \centering 
            \includegraphics[width=\textwidth]{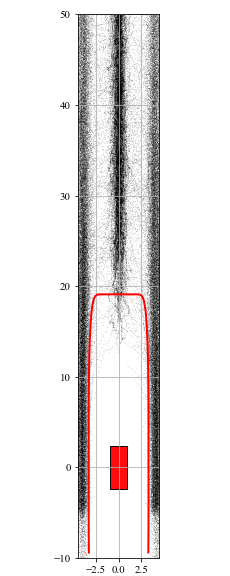}
            \caption[]%
            {{\small S-field at vehicle velocity 35 m$/$s}}    
            \label{fig:subj3}
        \end{subfigure}
        \caption {Visualization of S-field with parameters calibrated with highD data}
        \label{fig:subjective field}
    \end{figure}

Figure \ref{fig:subjective field} illustrates the S-field shapes corresponding to different velocities estimated upon highD vehicle data. The red contours around the red ego vehicle outline the estimated safety space it maintains, corresponding to the position where the S-field risk equals a threshold of $e^{-1}$. This threshold indicates a spatial proximity beyond which drivers' perceived proximity risk escalates significantly. 
The black dots overlaid in Figure \ref{fig:subjective field} represent the distribution of relative positions of surrounding vehicles, with each dot indicating the closest point on a surrounding vehicle relative to the ego vehicle.  

The estimated values for $\gamma_l$ and $\gamma_b$ are 1.18 and 1.64, respectively. This indicates that vehicles tend to maintain a greater lateral distance from lane markers compared to road boundaries. Additionally, the estimated values for $\beta_l$ and $\beta_b$ are 2.46 and 5.17, suggesting that vehicles are more sensitive to S-field risk when near road boundaries than when they are near lane markers.

\subsubsection{O-field parameters}

We choose the following set of parameters for the O-field: $\beta_d = 10$, $\beta_t=2$, and $\gamma_t=7.5$. The relatively large $\beta_d$ ensures the sensitivity of Eq. (\ref{eq:P}) to spatial proximity that potentially leads to collisions. The selection of $\gamma_t$ corresponds to the 5th-percentile value of positive TTC values in the dataset in ascending order, which is around 7.5 seconds. 

For $\gamma_d$, different values are assigned for different vehicle pairs to ensure proper detection of lateral collisions. We set $\gamma_d=0.5\times(w_i + w_j)$, where $w_i$ and $w_j$ denote the widths of the two vehicles. Since the vehicle angles in highD data are mostly below 3$\degree$, the most likely type of lateral collision is a side-swipe. The chosen method for calculating $\gamma_d$ could effectively capture potential side-swipe collisions while minimizing false alerts. In the longitudinal dimension, $\gamma_d$ is also capable of capturing rear-end collisions. 

\begin{figure}
        \centering
        \begin{subfigure}[H]{0.85\textwidth}
            \centering
            \includegraphics[width=\textwidth]{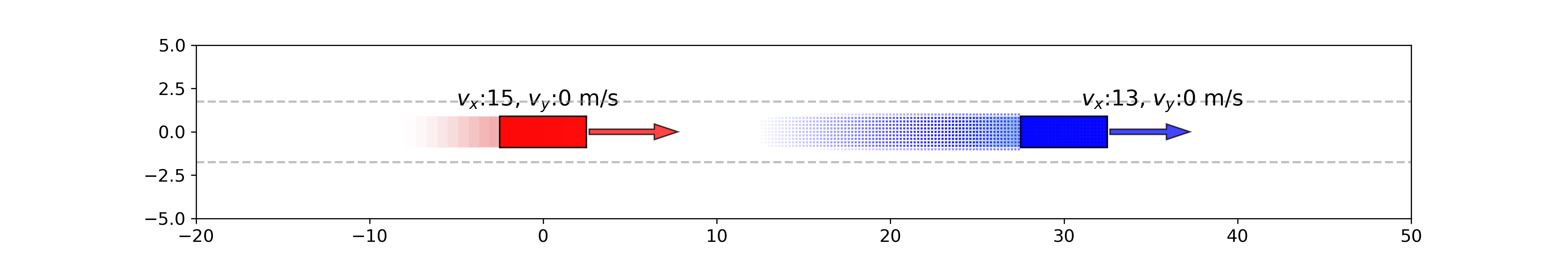}
            \caption[]%
            {{\small O-field in car-following scenario with a speed difference of 2 m$/$s}}    
            \label{fig:ob1}
        \end{subfigure}
        \begin{subfigure}[H]{0.85\textwidth}  
            \centering 
            \includegraphics[width=\textwidth]{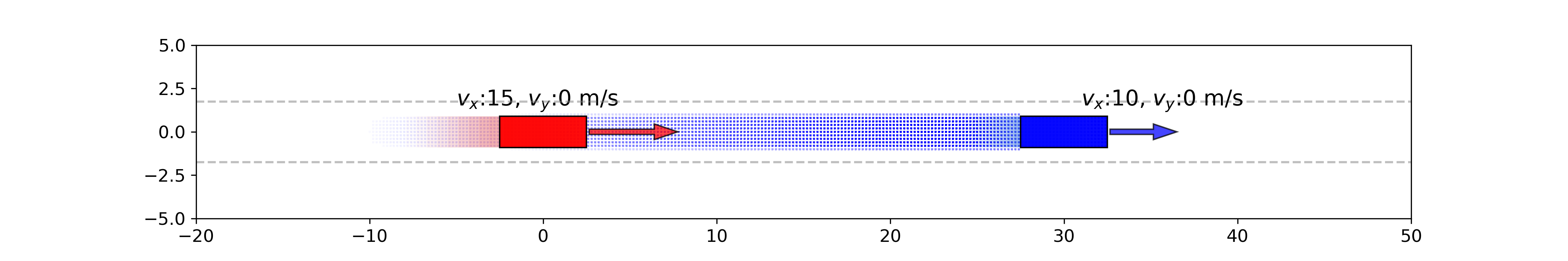}
            \caption[]%
            {{\small O-field in car-following scenario with a speed difference of 5 m$/$s}}    
            \label{fig:ob2}
        \end{subfigure}
        \begin{subfigure}[H]{0.85\textwidth}   
            \centering 
            \includegraphics[width=\textwidth]{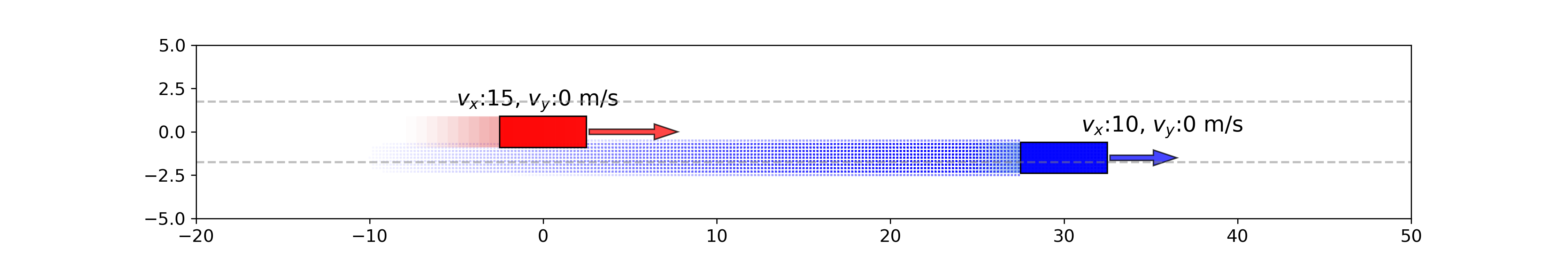}
            \caption[]%
            {{\small O-field in a potential side-swipe scenario}}    
            \label{fig:ob3}
        \end{subfigure}
        \begin{subfigure}[H]{0.85\textwidth}   
            \centering 
            \includegraphics[width=\textwidth]{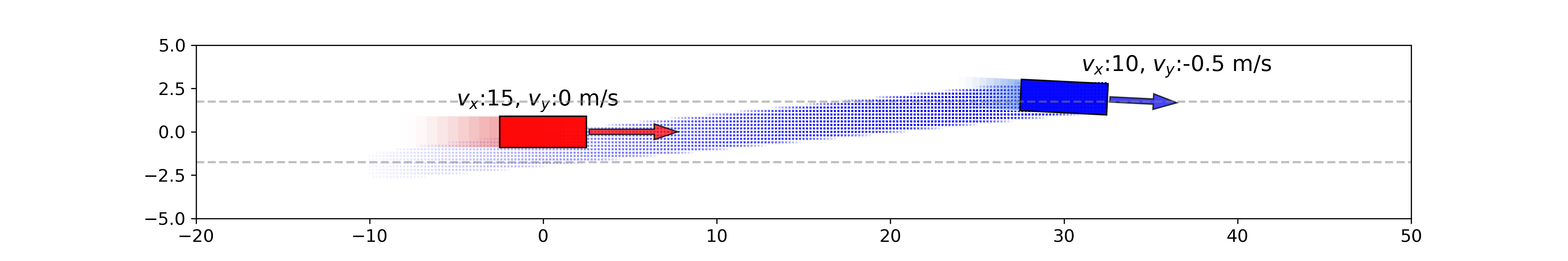}
            \caption[]%
            {{\small O-field in a lane-changing scenario with potential collision}}    
            \label{fig:ob4}
        \end{subfigure}
        \caption {Visualization of O-field with different inter-vehicle motions}
        \label{fig:objective field}
    \end{figure}

Figure \ref{fig:objective field} visualizes the shape of the O-field using the specified parameters under different scenarios of inter-vehicle motion. In all four scenarios depicted in Figure \ref{fig:objective field}, the ego vehicle is shown in red, maintaining a constant velocity of 15 m/s along its current lane. The position, velocity, and direction of the influencing vehicle, shown in blue, vary across scenarios to represent different relative motion conditions. The blue shaded areas indicate regions where the O-field risk exceeds a threshold of $e^{-1}$, consistent with the S-field threshold discussed earlier. When the ego vehicle is positioned inside these shaded areas, it is exposed to an elevated collision risk.

Figure \ref{fig:objective field}(a) and Figure \ref{fig:objective field}(b) depict the variations in the propagation of the O-field risk in car-following scenarios with different speed differences, with the latter extending further on the road, indicating a higher rear-end collision risk for the ego vehicle. Figure \ref{fig:objective field}(c) demonstrates the O-field's representation of a potential side-swipe scenario. Figure \ref{fig:objective field}(d) illustrates the O-field's characterization of potential collision in a lane-changing scenario. Overall, these demonstrated cases suggest that the O-field with the selected parameters effectively captures collision risks arising from both longitudinal and lateral inter-vehicle motions.

\subsection{Driving behavior analysis} \label{subsec: analysis}
This subsection validates the relationship between the proposed C-SPF and driver behaviors by applying the C-SPF to the highD dataset. The distributions of vehicle velocity and acceleration associated with risky scenarios identified by the subjective or objective components of the C-SPF are presented. The analysis found the C-SPF could effectively indicate drivers' braking and lateral maneuvers.

\subsubsection{Braking}

\begin{figure}[h!]
\centering
	\includegraphics[width=.85\textwidth]{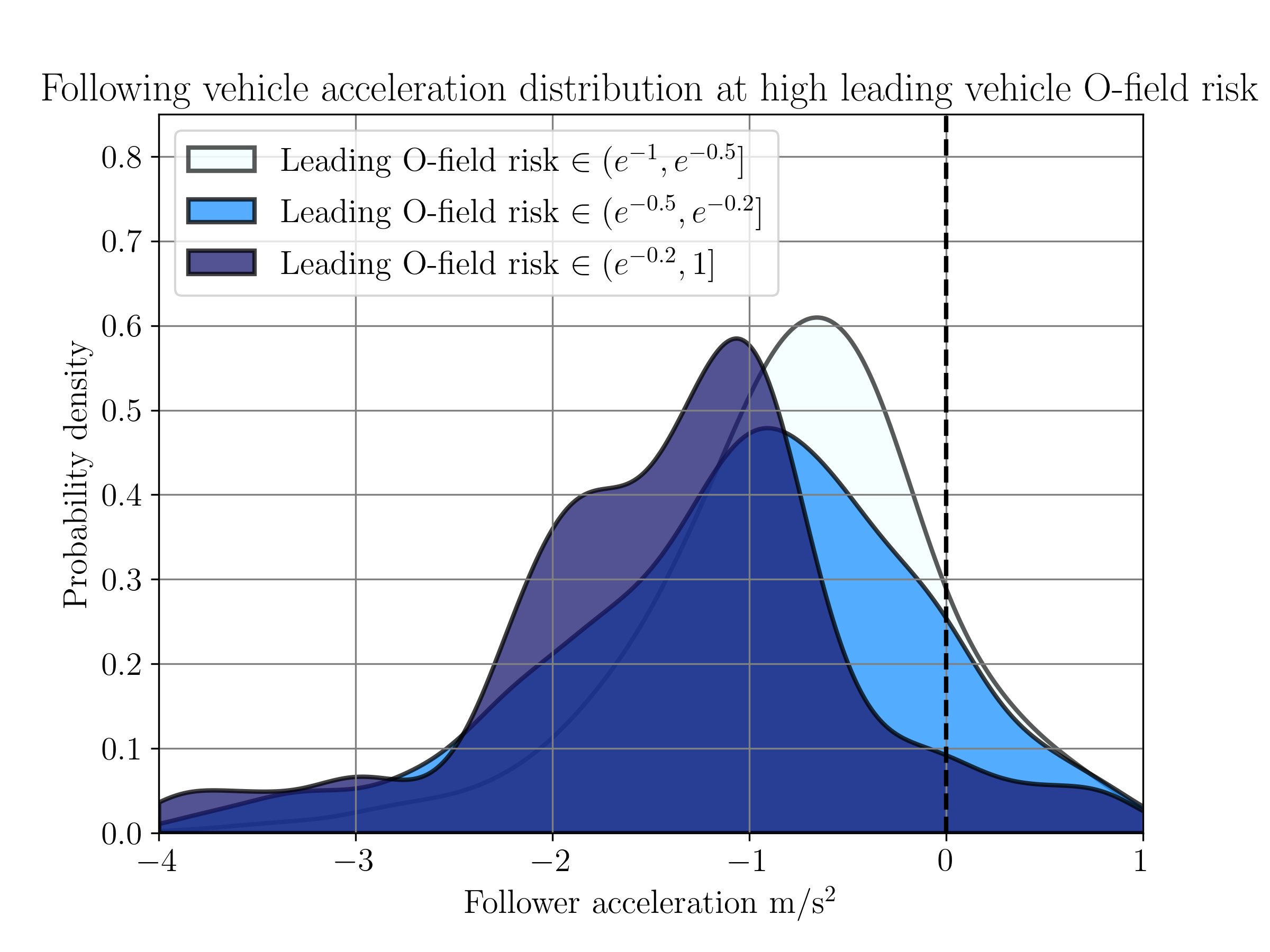}
	\caption{Following vehicle acceleration distribution 1 second after leading vehicle O-field risk peaking over threshold}
\label{fig:car-following distribution}
\end{figure}

In car-following scenarios, the following vehicle will apply the brakes upon perceiving a potential collision with the leading vehicle. The intensity of braking generally depends on the proximity to the potential collision. In the highD data, the braking behavior of following vehicles is observed to be associated with high O-field risk caused by the leading vehicle. Figure \ref{fig:car-following distribution} illustrates the distribution of the following vehicles' acceleration rates within 1 second after the leading vehicle posing an O-field risk exceeding certain thresholds, indicating a risk of rear-end collision. The distribution indicates that vehicles are likely to decelerate more when the O-field risk posed by the leading vehicle is high. Specifically, greater deceleration is associated with higher levels of O-field risk from the vehicle ahead. This demonstrates that the O-field in the proposed C-SPF effectively captures rear-end collision risks that trigger drivers' braking behaviors.

\subsubsection{Lateral maneuver}

\begin{figure}[h!]
\centering
	\includegraphics[width=.85\textwidth]{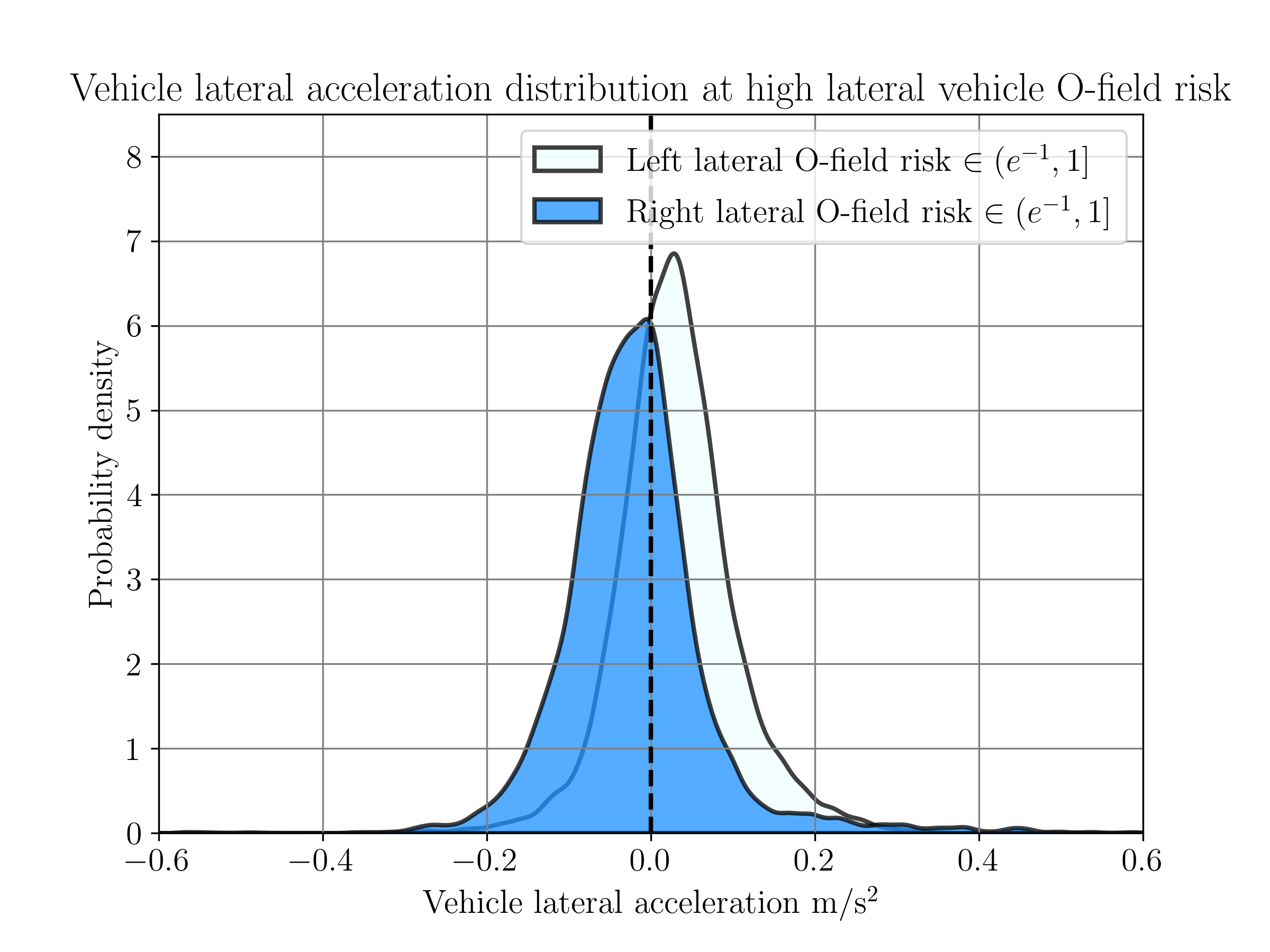}
	\caption{Distribution of vehicle lateral acceleration within 1 second after the lateral O-field risk exceeds a threshold. Lane-changing vehicles are isolated. Rightward acceleration is considered positive, while leftward acceleration is negative. }
\label{fig:lateral acceleration o-risk}
\end{figure}

\begin{figure}[h!]
\centering
	\includegraphics[width=.85\textwidth]{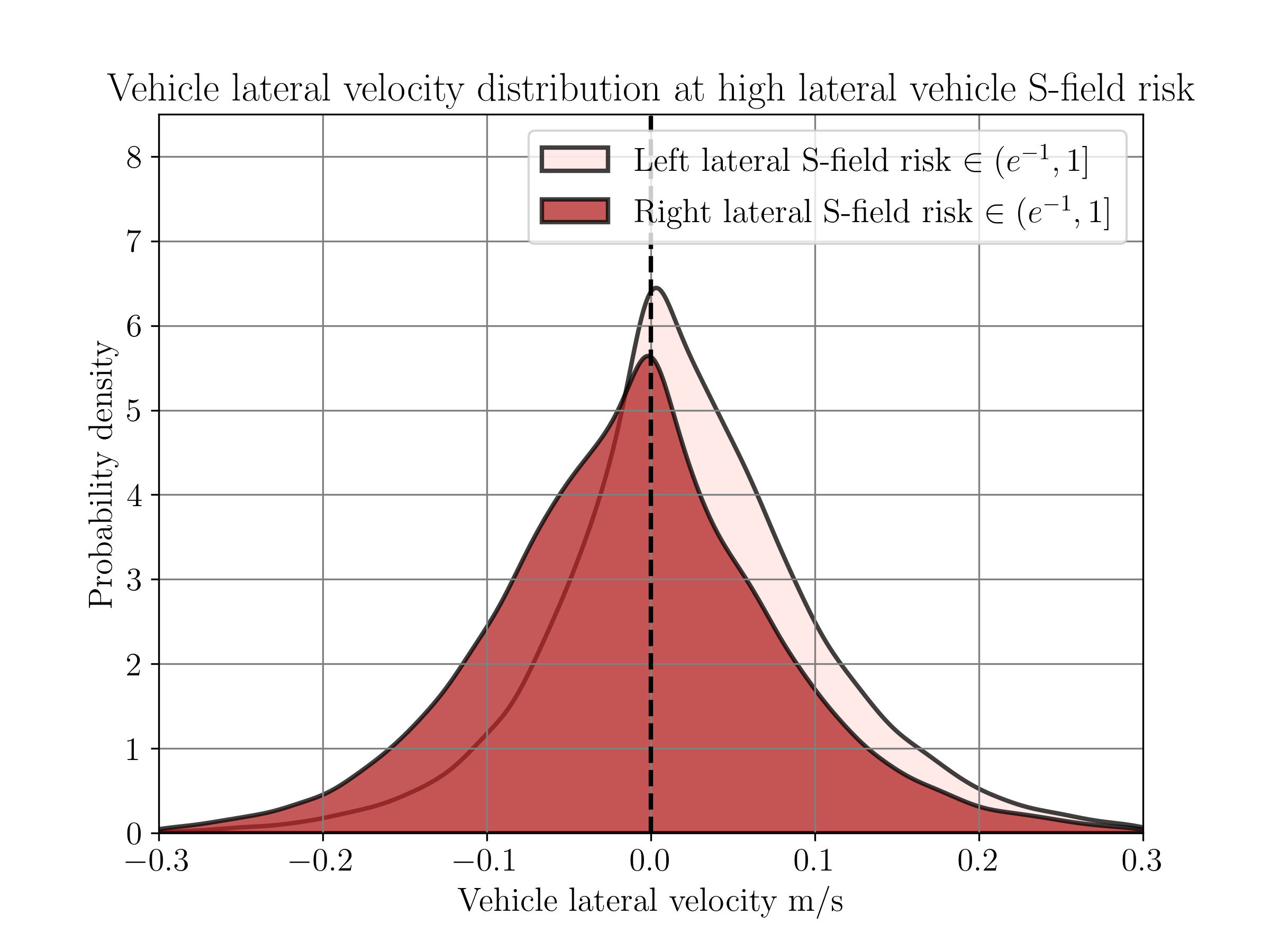}
	\caption{Distribution of vehicle lateral velocity within 1 second after the lateral S-field risk exceeds a threshold. Lane-changing vehicles are isolated. Rightward velocity is considered positive, while leftward velocity is negative.}
\label{fig:lateral velocity s-risk}
\end{figure}

C-SPF is capable of capturing risks that influence vehicles' maneuvers in the lateral dimension. Excluding lane-changing vehicles from the dataset, Figure \ref{fig:lateral acceleration o-risk} shows the distribution of vehicles' lateral acceleration within 1 second after vehicles in adjacent lanes generate a relatively high O-field risk, indicating an elevated collision risk from the lateral. The comparison of vehicles' lateral acceleration rates under the influences of left and right lateral O-field risks shows that vehicles accelerate laterally in the direction opposite to the source of the risk. For example, when a vehicle in the left adjacent lane produces a relatively high O-field risk, the influenced vehicle tends to accelerate toward the right. 

Figure \ref{fig:lateral velocity s-risk} illustrates the distribution of vehicles' lateral velocity within 1 second after being influenced by a relatively high S-field risk from the adjacent lanes. Lane-changing vehicles are excluded. The figure demonstrates that the distribution of lateral velocities skews towards the direction opposite to the source of the lateral proximity risk. This indicates that vehicles tend to adjust their lateral position in the lane when perceiving an escalated proximity risk from the lateral, causing driver discomfort and encouraging them to maneuver.


\subsection{Case studies}\label{subsec:case studies}

To validate the ability of the proposed C-SPF risk indicators to identify driving risks and interpret driver behaviors across multiple scenarios, three cases randomly selected from the highD data are discussed, including: (1) a car-following scenario where the following brake to avoid rear-end collisions with the leading vehicle in a stop-and-go traffic (2) a lateral risky scenario where a car gave up lane-change due to the potential collision risk the preceding vehicle in the target lane (3) a passing scenario where the vehicle adjust its lateral position in the lane to ensure acceptable spacing with vehicle in the adjacent lane. The S-field risk associated with lane markers and boundaries is not considered in case studies. Two baseline models are compared with the C-SPF risk indicators:
\begin{itemize}
\item Inverse time-to-collision (TTCi): This metric is calculated as the reciprocal of the inter-vehicle TTC determined in two dimensions. The two-dimensional algorithm from (\cite{hou2015new}) is used to compute TTC by numerically simulating the movement of vehicles' rectangular bounding boxes along their current trajectories and speeds until a potential collision occurs. The time until this collision is then recorded as the TTC estimated at the current time point.
\item Relative driving safety index (RDSI) (\cite{wang2016driving}): a safety potential field-based risk index deriving driving risks from a psychological field incorporating both subjective and objective risk factors to a risk function. RDSI quantifies both the potential energy of the vehicle within the field and the rate of change of this potential energy over time. The parameters specified in the original paper of \cite{wang2016driving} are used for its implementation.
\end{itemize}

\begin{figure}
        \centering
        \begin{subfigure}[H]{0.95\textwidth}
            \centering
            \includegraphics[width=\textwidth]{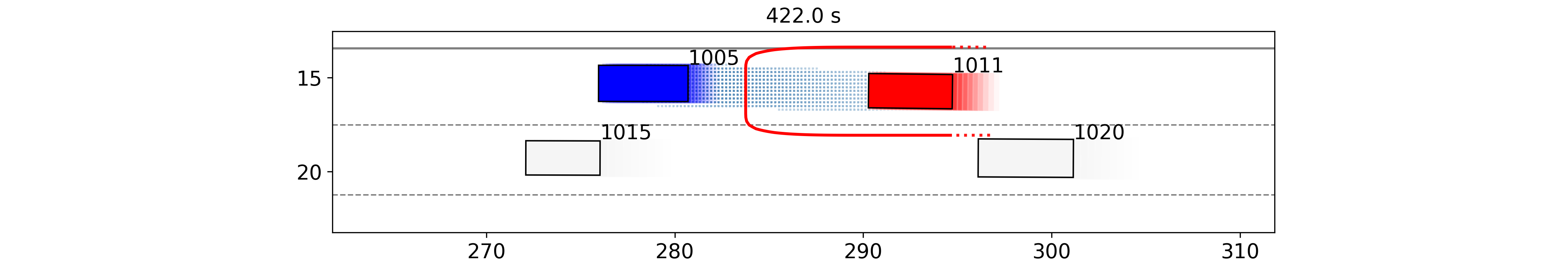}
            \caption[Network2]%
            {{\small Car-following case 422.0 s}}    
            \label{fig:cf1}
        \end{subfigure}
        \hfill
        \begin{subfigure}[H]{0.95\textwidth}  
            \centering 
            \includegraphics[width=\textwidth]{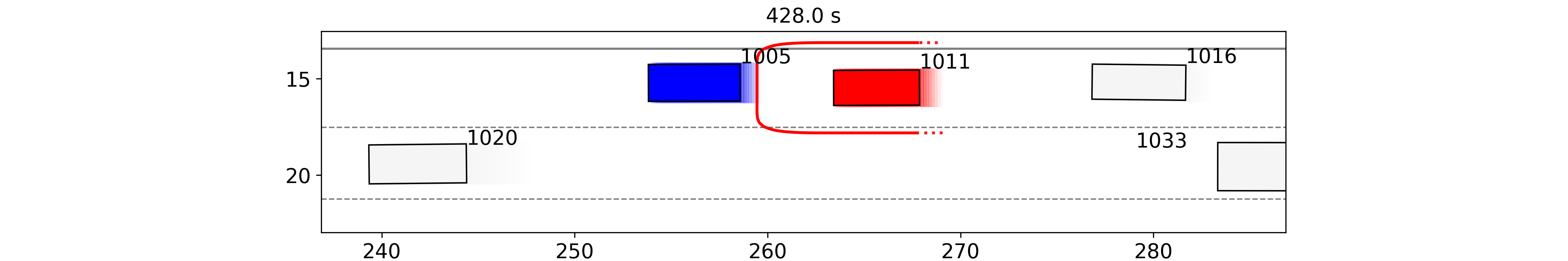}
            \caption[]%
            {{\small Car-following case 428.0 s}}    
            \label{fig:cf2}
        \end{subfigure}
        \begin{subfigure}[H]{0.95\textwidth}   
            \centering 
            \includegraphics[width=\textwidth]{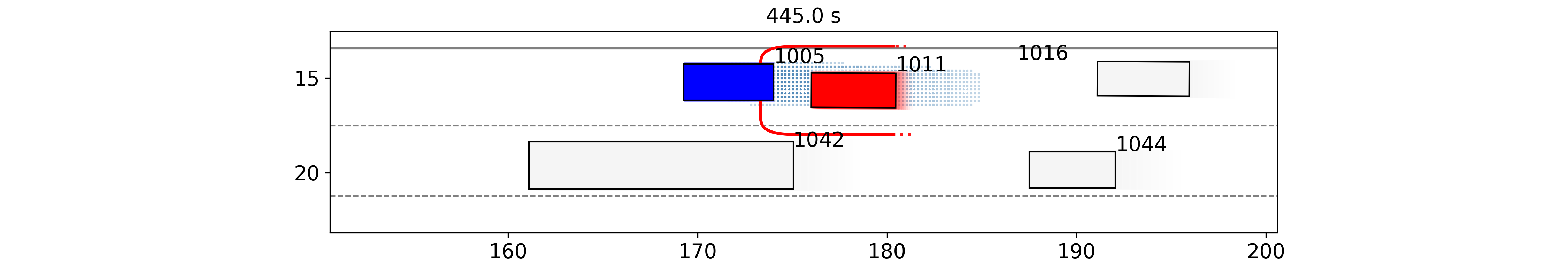}
            \caption[]%
            {{\small Car-following case 445.0 s}}    
            \label{fig:cf3}
        \end{subfigure}
        \caption {Visualization of car-following case. Vehicle No.1011 (red) is the ego vehicle which is the following vehicle in car-following and vehicle No.1005 (blue) is the leading vehicle. The red contour represents the ego vehicle's subjective safety space; when the leading vehicle intrudes into this contour, the S-field risk exceeds $e^{-1}$. The blue shaded areas indicate regions where the leading vehicle poses an O-field risk greater than $e^{-1}$.}
        \label{fig:car-following visual}
    \end{figure}

\begin{figure}[!h]
\centering
	\includegraphics[width=.85\textwidth]{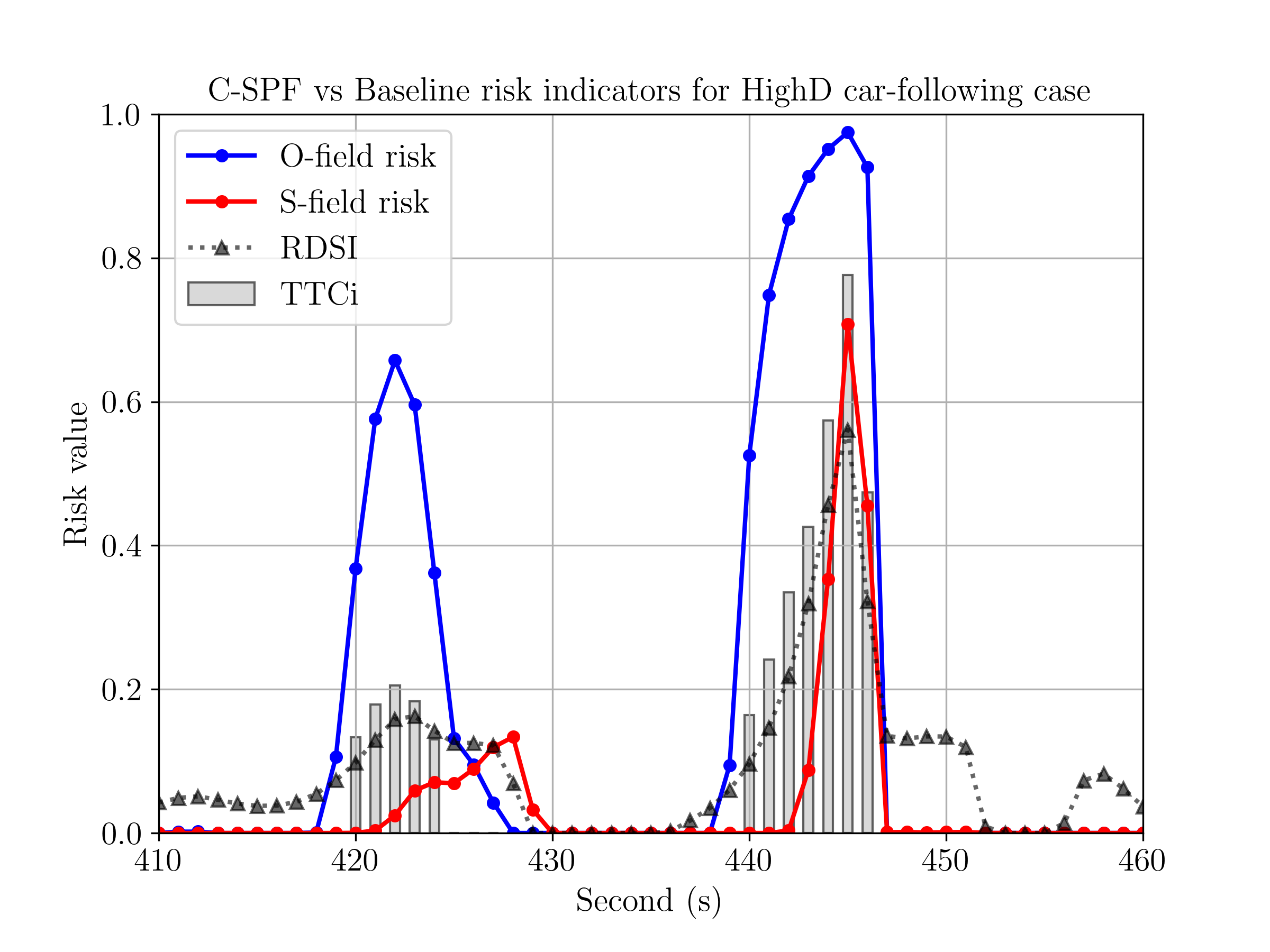}
	\caption{Comparison of the change of C-SPF and baseline risks experienced by the following vehicle in the car-following case}
\label{fig:car-following case}
\end{figure}

\subsubsection{Car-following}
Figure \ref{fig:car-following visual} illustrates a risky car-following scenario in the highD dataset. The ego vehicle No.1011, in red color, followed by the leading vehicle No.1005, in blue color. This is a stop-and-go traffic scenario. The leading vehicle stopped twice at around time 420.0 s and 440.0 s. The ego vehicle No.1011, which is the following vehicle of car-following, applied brakes to avoid a rear-end collision. Figure \ref{fig:car-following case} illustrates the change of C-SPF and baseline risks experienced by the following vehicle in the car-following case. The ego vehicle, which was the follower in this longitudinal interaction, was exposed to elevated O-field risk from the leading vehicle when approaching the leading vehicle between 420.0 s and 425.0 s and between 440.0 s and 445.0 s. These high O-field risk values originate from the longitudinal speed difference between the ego vehicle and the leading vheicle, indicating a high rear-end collision risk with respect to the leading vehicle. 
In Figures \ref{fig:car-following visual} and \ref{fig:car-following visual}, the regions where the O-field risk posed by the leading vehicle exceeds a threshold of $e^{-1}$ are highlighted with blue shading. A threshold of $e^{-1}$ is chosen for visualizing the O-field because, as shown in Figure \ref{fig:car-following distribution}, an O-field risk greater than $e^{-1}$ is sufficient to trigger vehicle deceleration.

A small S-field risk increase was observed at 428.0 s, indicating a slight increase in proximity risk due to reduced spacing with the leading vehicle following braking. A more pronounced S-field risk peak occurred around 445.0 s, triggered by the diminishing distance to the leading vehicle. As shown in Figure \ref{fig:cf3}, the leading vehicle (depicted in blue) intruded into the ego vehicle's estimated safety space, resulting in an elevated S-field risk. 

As for comparison with the baseline safety indicators, \ref{fig:car-following case} demonstrates that the O-field mostly aligned with the trend of TTCi but is more sensitive to imminent rear-end collisions. RDSI also captured the potential rear-end collision risk between 420.0 s and 425.0 s and between 440.0 s and 445.0 s. However, this model exhibited more fluctuations compared to TTCi and the O-field, which may reduce its reliability in consistently identifying high-risk situations.

\subsubsection {Lane-change give-up}

\begin{figure*}[h!]
        \centering
        \begin{subfigure}[H]{0.95\textwidth}  
            \centering 
            \includegraphics[width=\textwidth]{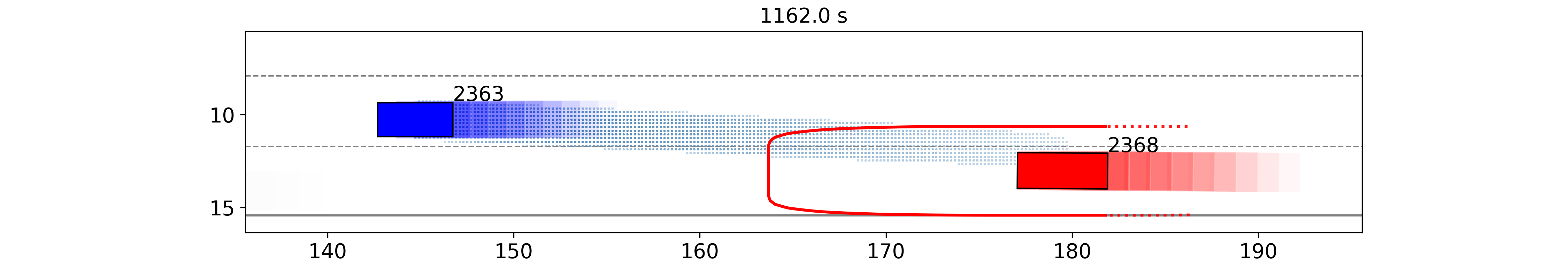}
            \caption[]%
            {{\small Lane-change give-up case 1162.0 s}}    
            \label{fig:lc-giveup2}
        \end{subfigure}
        \begin{subfigure}[H]{0.95\textwidth}   
            \centering 
            \includegraphics[width=\textwidth]{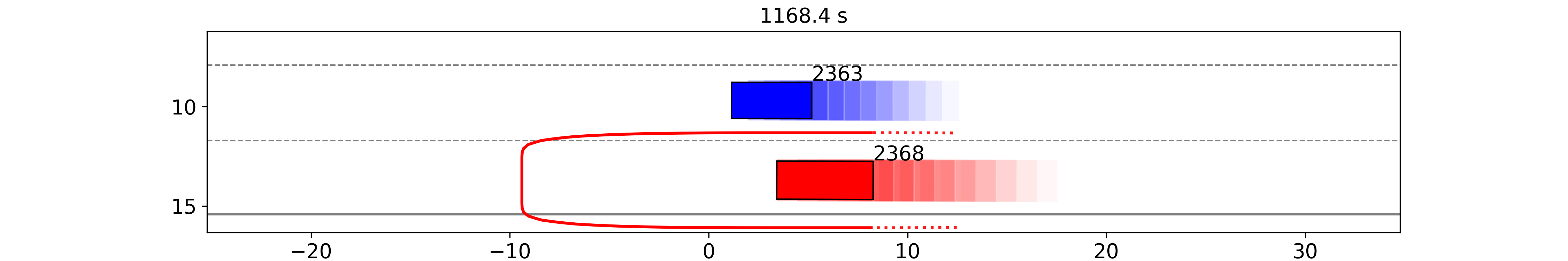}
            \caption[]%
            {{\small Lane-change give-up case 1168.4 s}}    
            \label{fig:lc-giveup3}
        \end{subfigure}
        \caption {Visualization of a scenario where Vehicle No. 948 adjusted its lateral position in response to Vehicle No. 944 in the adjacent right lane, which deviated from the center of its lane. The red contour represents the ego vehicle's subjective safety space; when the other vehicle intrudes into this contour, the S-field risk exceeds $e^{-1}$. The blue shaded areas indicate regions where the other vehicle poses an O-field risk greater than $e^{-1}$.}
        \label{fig:lc-giveup visual}
    \end{figure*}

\begin{figure}
\centering
\includegraphics[width=.85\textwidth]{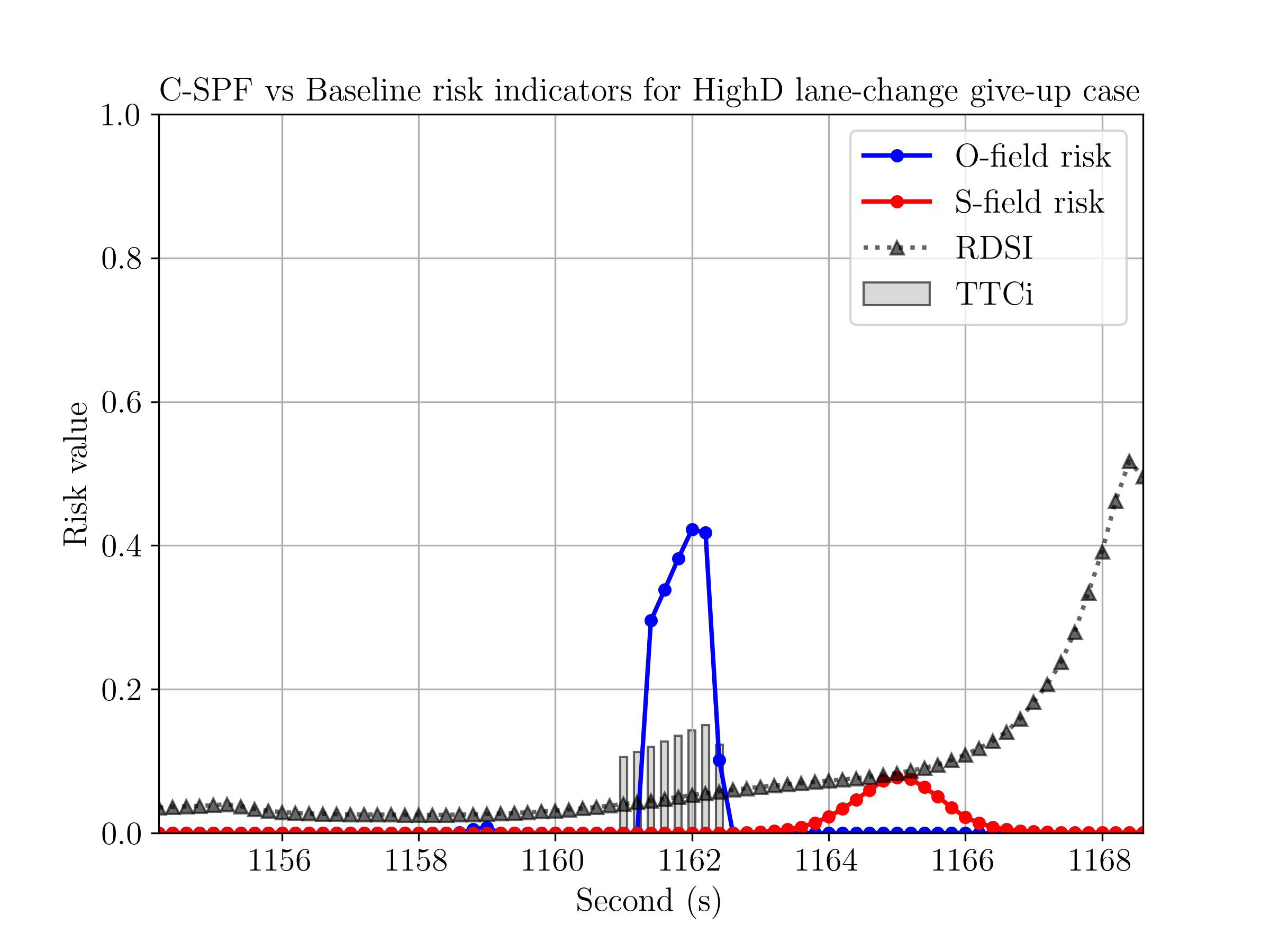}
	\caption{Comparison of the change of C-SPF and baseline risks experienced by ego vehicle No. 2368 who gave up the lane-change}
\label{fig:lc give-up case}
\end{figure}
Figure \ref{fig:lc-giveup visual} presents a case study where the ego vehicle No. 2368 abandoned a lane-change attempt due to the potential collision risk posed by the preceding vehicle No. 2363 in the target lane. The risks experienced by the ego vehicle, assessed using the proposed C-SPF and two baseline metrics, are depicted in Figure \ref{fig:lc give-up case}. From 1160.0 s to 1162.0 s, ego vehicle No. 2368 was preparing for a lane change. As shown in Figure \ref{fig:lc-giveup visual}(b), the ego vehicle began to depart from its current lane at 1162.0 s but decided to abort the lane-change maneuver upon recognizing the potential collision risk with vehicle No. 2363 in the target lane. Subsequently, the ego vehicle returned fully to its current lane and continued driving in it.

The C-SPF O-field detected a potential collision risk with the preceding vehicle No. 2363 in the target lane at 1162.0 s, as shown in Figure \ref{fig:lc give-up case}. Additionally, the blue-shaded area in Figure \ref{fig:lc-giveup visual}(b) delineates the C-SPF O-field generated by vehicle No. 2363 on the ego vehicle, with the shaded region indicating O-field risk values exceeding a threshold of $e^{-1}$. This threshold is used for visualization because, as discussed in Section \ref{subsec: analysis}, a lateral O-field risk above $e^{-1}$ can trigger vehicles to perform lateral maneuvers to mitigate the risk. The O-field risk values between 1161.0 s and 1162.0 s align with the calculations of the two-dimensional TTCi, with both measures identifying the potential collision risk if the ego vehicle continued on its current course.

In contrast, RDSI failed to capture this potential collision risk triggering the ego vehicle to abort the lane-change maneuver. Instead, it flagged the moment at 1168.4 s when the two vehicles were side by side while passing. Furthermore, a slightly elevated C-SPF S-field risk was observed due to the short relative spacing to the right preceding vehicle No. 2368 before vehicle No. 2363 fully returned to its current lane. The comparison suggests the proposed C-SPF is able to interpret drivers' intention to abandon lane-changing due to potential lateral collision risks.

\subsubsection{Lateral position adjustment}
\begin{figure*}
        \centering
        \begin{subfigure}[H]{0.95\textwidth}
            \centering
            \includegraphics[width=\textwidth]{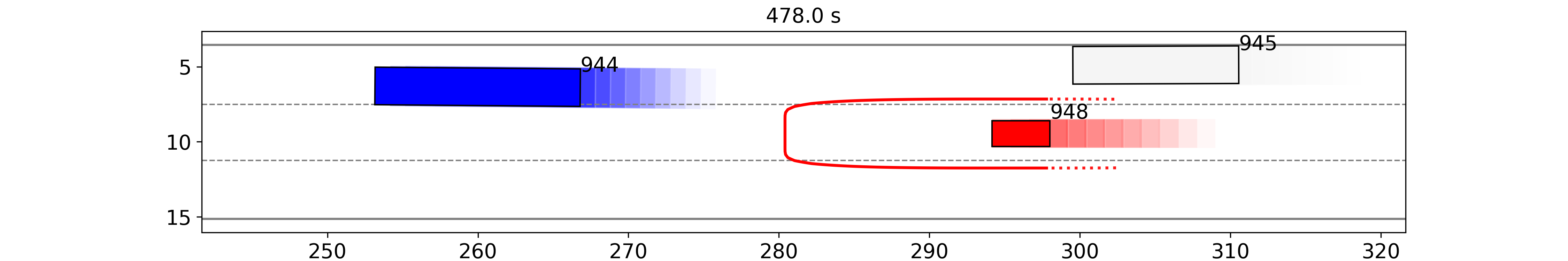}
            \caption[Network2]%
            {{\small Lateral position adjustment case 478.0 s}}    
            \label{fig:lateral1}
        \end{subfigure}
        \hfill
        \begin{subfigure}[H]{0.95\textwidth}  
            \centering 
            \includegraphics[width=\textwidth]{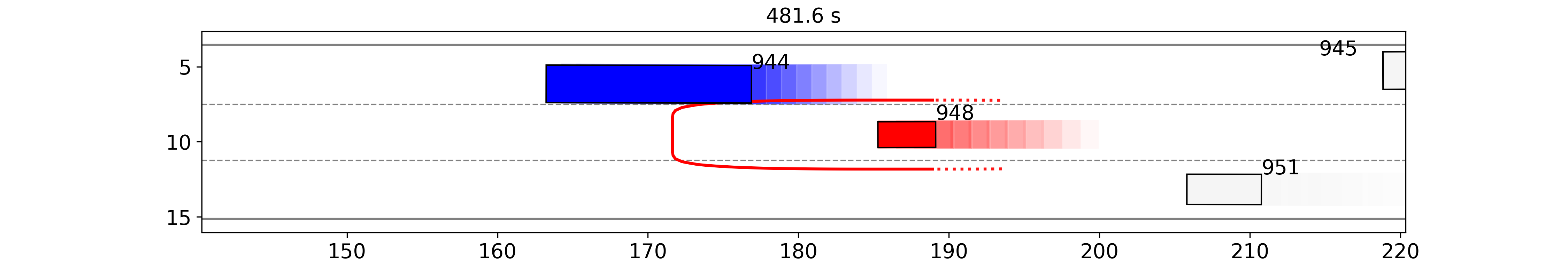}
            \caption[]%
            {{\small Lateral position adjustment case 481.6 s}}    
            \label{fig:lateral2}
        \end{subfigure}
        \begin{subfigure}[H]{0.95\textwidth}   
            \centering 
            \includegraphics[width=\textwidth]{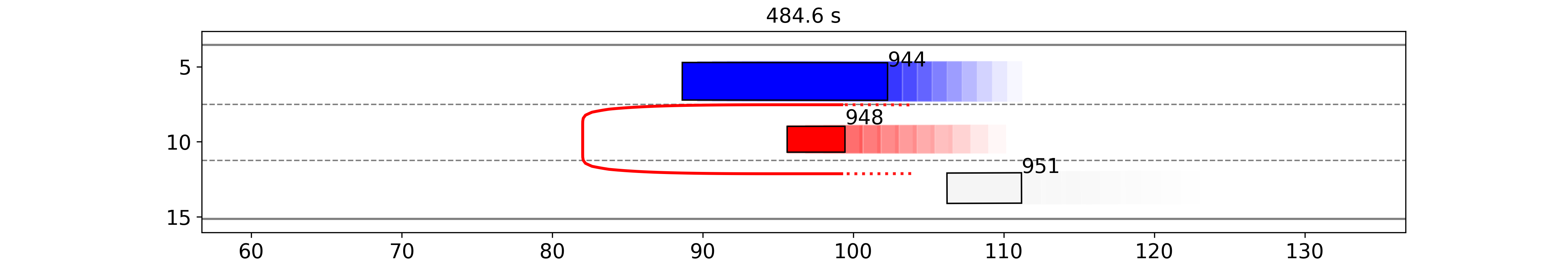}
            \caption[]%
            {{\small Lateral position adjustment case 484.6 s}}    
            \label{fig:lateral3}
        \end{subfigure}
        \caption {Visualization of a scenario where Vehicle No. 948 adjusted its lateral position in response to Vehicle No. 944 in the adjacent right lane, which deviated from the center of its lane. The red contour represents the ego vehicle's subjective safety space; when the other vehicle intrudes into this contour, the S-field risk exceeds $e^{-1}$. The blue shaded areas indicate regions where the other vehicle poses an O-field risk greater than $e^{-1}$.}
        \label{fig:lateral position visual}
    \end{figure*}

\begin{figure}
\centering
\includegraphics[width=.85\textwidth]{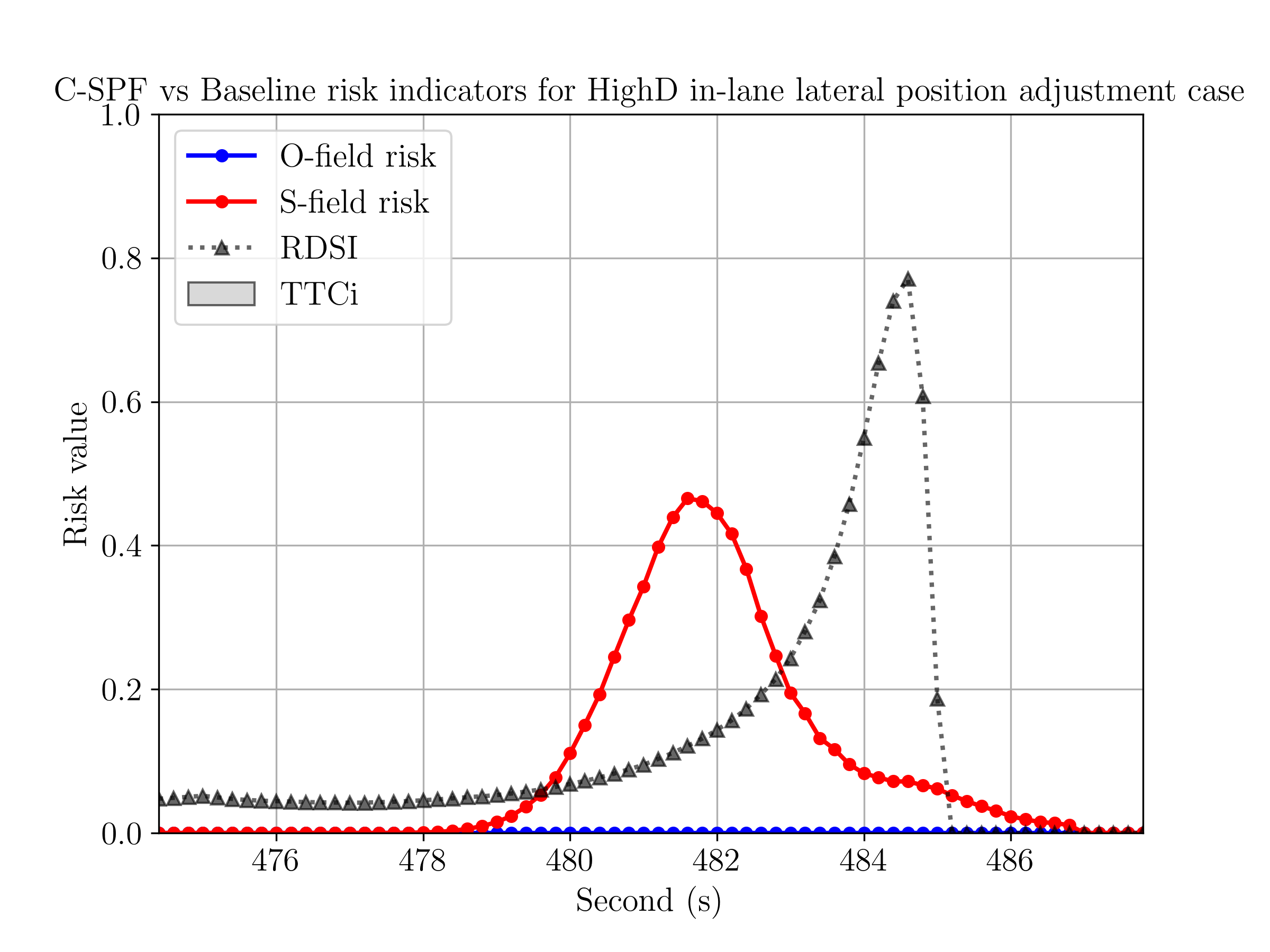}
	\caption{Comparison of the change of C-SPF and baseline risks experienced by ego vehicle No. 948 in the lateral adjustment case}
\label{fig:lateral position case}
\end{figure}

Figure \ref{fig:lateral position visual} illustrates a scenario where the ego vehicle No. 948 passed another vehicle, No. 944, in the right adjacent lane. Vehicle No. 944 deviated from its lane, reducing the lateral spacing with No. 948. This reduction triggered a peak in the S-field risk around 481.6 s, as shown in Figure \ref{fig:lateral position case}. In response to this risk, the ego vehicle No. 948 steered toward the left side of its current lane while passing to mitigate the perceived proximity risk. Figure \ref{fig:lateral position visual}(b) shows the situation before the lateral adjustment, where the lateral vehicle intruded into the safety space estimated for the ego vehicle. Figure \ref{fig:lateral position visual}(c) depicts the moment after the lateral adjustment, where No. 948 had safely passed vehicle No. 944 without significant proximity risk from the lateral.

During the entire process of passing and adjusting the lateral position, neither the TTCi nor the O-field risk detected any tendency for a collision. This suggests that the lateral maneuver of ego vehicle No. 948 was not prompted by a potential collision risk. Instead, it was influenced by the driver’s perception of a proximity risk from vehicle No. 944 in the right lane, which was effectively captured by the S-field component in the proposed C-SPF, as shown in Figure \ref{fig:lateral position visual}(b). In contrast, as shown in Figure \ref{fig:lateral position case}, RDSI failed to alert the lateral proximity risk before the maneuver and only flagged a risk at the moment of passing, around 484.6 s. This highlights the superior interpretive power of the proposed C-SPF in understanding drivers' lateral maneuvers compared to this baseline safety potential field model.

\section{Discussion}\label{sec:discussion}

The above analysis suggests that the proposed C-SPF effectively captures both collision and non-collision risks across various scenarios. These case studies demonstrate the proposed C-SPF has better interpretive power to driver behaviors in various scenarios, such as braking, lane-change abortion, and lateral position adjustment, compared to existing models. This can be attributed to three advantages of C-SPF over these models.

First, with its composite structure, the C-SPF offers a comprehensive risk assessment metric by integrating drivers' subjective risk perceptions regarding proximity and the objective risk of potential collisions. The proximity risk perception and the collision risk are often independent as demonstrated in Section \ref{subsec:case studies}, where escalated lateral proximity risks triggered drivers' maneuvering in situations without a rising collision risk. The C-SPF using separate risk components can better capture and generalize both risk dimensions compared to the baseline safety potential field (\cite{wang2016driving}), which attempts to unify both risks within a single psychological field.

Second, compared to other safety potential field models, the subjective component of the C-SPF can be calibrated using abundant spacing data derived from publicly available highway segment trajectory datasets. Additionally, the objective component of the C-SPF relies on straightforward equations of vehicle motion to predict collisions, eliminating the need for complex parameters and ensuring ease of implementation. Existing safety potential field models, such as RDSI, are often insufficiently calibrated, relying on arbitrary physical functions and requiring local accident statistics, which may not be available for every highway segment. Other models in the literature propose calibrating safety potential fields based on drivers' deceleration behaviors in car-following scenarios. However, this approach provides only longitudinal information, limiting the model's ability to explain lateral maneuvers. In contrast, spacing data is two-dimensional, enabling the C-SPF to incorporate both longitudinal and lateral information, providing a more comprehensive understanding of drivers' risk perception.

The C-SPF offers significant potential for applications in pre-collision warning systems. A warning signal can be triggered when the risk level surpasses a predefined threshold, alerting drivers to imminent dangers. Additionally, the C-SPF can be seamlessly integrated into ADAS or autonomous driving systems to improve their understanding of surrounding risks. By leveraging outputs from both the S-field and O-field, these systems can potentially be more sensitive to both longitudinal and lateral risks and achieve a more human-like risk assessment, enhancing their ability to interact safely and intuitively with other vehicles in traffic.

\section{Conclusion}\label{sec:conclusion}

In this study, we propose a new composite safety potential field (C-SPF) framework to comprehensively assess highway driving risks and interpret driver behavior. The C-SPF integrates a Subjective Field (S-field) to estimate drivers' perceptions of spatial proximity risks to surrounding entities and an Objective Field (O-field) to predict collision probabilities between vehicles. Compared to existing safety potential field models, the C-SPF avoids reliance on the sparsely available data of accidents and is simpler to calibrate with abundant two-dimensional vehicle spacing data in publicly available trajectory datasets. With its composite structure, the C-SPF can both mimic human drivers' risk perception of spatial proximity and objectively assess the likelihood of potential road collisions.

The analysis and case studies demonstrate that the proposed C-SPF provides superior explainability for risky situations and the corresponding driver maneuvers compared to existing risk indicators, such as TTC and RDSI. The experiments driven by naturalistic data reveal that the C-SPF effectively identifies risks that trigger drivers to brake or perform lateral maneuvers. Additionally, the case studies highlight the C-SPF's ability to identify risk sources that trigger drivers' safety maneuvers, which existing indicators either fail to capture or detect with delay, particularly in scenarios involving lateral interactions.

However, our current study has several limitations. First, only absolute speed is considered in shaping the S-field due to data limitations and the need for simpler calibration. Other factors, such as road curvature, ramps, and vehicle class, can also influence drivers' perception of proximity risks. Future research should explore more effective calibration methods to incorporate these factors into the model. Second, for simplicity, the calculation of O-field risk assumes constant vehicle speeds, and the spatial component of the O-field does not account for vehicle length. While the current implementation of the O-field is efficient, it could be improved by incorporating speed variations and more detailed vehicle geometries for greater accuracy.

Future research will focus on several key areas. First, the model will be improved by incorporating road and vehicle factors in the S-field, as well as accounting for speed variations and vehicle geometries to improve the accuracy of the O-field. Second, the interconnection between the S-field and O-field will be explored to better understand the nature of driving risk on highways. Third, further experiments will explore the potential of utilizing C-SPF to enhance vehicle safety applications, such as collision warning systems and ADAS, by providing human-like risk assessment and improving collision risk avoidance.

\bibliographystyle{plainnat}
\bibliography{main}

\end{document}